%% file: emnlp2022.tex
\pdfoutput=1

\documentclass[11pt]{article}

\usepackage[]{EMNLP2022}

\usepackage{times}
\usepackage{latexsym}

\usepackage[T1]{fontenc}

\usepackage[utf8]{inputenc}

\usepackage{microtype}

\usepackage{inconsolata}

\usepackage{booktabs, colortbl} 
\usepackage{multirow,makecell}
\usepackage{multicol}
\usepackage{amsmath}
\usepackage{amssymb}
\usepackage{caption}
\usepackage{subfig}
\usepackage{titlesec}
\usepackage{pgfplots}
\usepackage[normalem]{ulem} 
\usepackage{enumitem}
\usepackage{graphicx}
\usepackage{bbding}
\usepackage{hyperref}

\usepackage{diagbox}

\usepackage{enumitem}

\usepackage{pifont}

\usepackage{amsmath}

\DeclareMathOperator*{\argmin}{arg\,min}

\title{Learning an Artificial Language for Knowledge-Sharing \\in Multilingual Translation}

\author{Danni Liu \and Jan Niehues \\
        Karlsruhe Institute of Technology \\
        \texttt{\{danni.liu, jan.niehues\}@kit.edu}} 

\begin{document}
\maketitle
\hyphenation{spBLEU}

\begin{abstract}
The cornerstone of multilingual neural translation is shared representations across languages.
Given the theoretically infinite representation power of neural networks, semantically identical sentences are likely represented differently.
While representing sentences in the \textit{continuous} latent space ensures expressiveness, 
it introduces the risk of capturing of irrelevant features which hinders the learning of a common representation.
In this work, 
we \textit{discretize} the encoder output latent space of multilingual models by assigning encoder states to entries in a codebook,
which in effect represents source sentences in a new artificial language.
This discretization process not only offers a new way to interpret the otherwise black-box model representations,
but, more importantly, gives potential for increasing robustness in unseen testing conditions.
We validate our approach on large-scale experiments with realistic data volumes and domains.
When tested in zero-shot conditions, 
our approach is competitive with two strong alternatives from the literature.
We also use the learned artificial language to analyze model behavior, 
and discover that using a similar bridge language increases knowledge-sharing among the remaining languages.\footnote{Code available at: \url{https://github.com/dannigt/fairseq/tree/master/examples/quant}}
\end{abstract}

\section{Introduction} \label{sec:intro}
A promising potential of multilingual~\cite{dong-etal-2015-multi,firat-etal-2016-multi,ha-etal-2016-toward,johnson-etal-2017-googles} neural machine translation (NMT) is knowledge-sharing between languages.
To enable knowledge-sharing, a prerequisite is the ability to capture common features of languages, especially between related ones.
\textit{Constructed languages} such as \textit{Interlingua} and \textit{Esperanto} are excellent examples of human-designed structures based on the commonalities of a wide range of related languages.
For data-driven models, however, it is difficult to leverage such resources due to data scarcity:
There is little parallel data to these constructed languages, 
and creating new translation heavily depends on expert curation.
Instead of relying on manually-created data, 
we aim to learn an artificial language in a more unsupervised fashion in parallel with training the NMT model.
Specifically,
our goal is to learn a sequence of tokens to represent the source sentences,
which then serves as context for the NMT decoder. 
\autoref{tab:codes-example} illustrates this  idea.

\begin{table}[]
\setlength\tabcolsep{3pt} 
    \centering
    \small
    \begin{tabular}{r|ccccc}
         \textbf{source sentence}
         & \textcolor{purple}{learning}
         & a 
         & \textcolor{brown}{new}
         & \textcolor{blue}{language} \\
         (English) & $\downarrow$ & $\downarrow$  & $\downarrow$  & $\downarrow$\\
         \textbf{discrete codes}
         & 3 & 609 & 57 & 1042 \\
         \\
         \\
         \textbf{source sentence}
         & \textcolor{purple}{belajar}
         & \textcolor{blue}{bahasa}
         & \textcolor{brown}{baru} \\
         
         (Indonesian) & $\downarrow$ & $\downarrow$  & $\downarrow$ \\
         \textbf{discrete codes}
         & 3 & 57 & 258 \\
    \end{tabular}
    \caption{We aim to learn a sequence of discrete codes to represent source sentences in multilingual NMT models. 
    Our goal is to 1) improve inference-time robustness, 2) have more interpretable intermediate representations.
    }
    \label{tab:codes-example}
\end{table}


A potential advantage of representing inputs in discrete tokens is \textit{robustness}, 
a property especially relevant when NMT systems must cope with unexpected testing conditions.
By discretization, we restrict the continuous latent space to a finite size, 
providing the possibility for model intermediate representations to fall back to a position seen in training. 
For instance, in zero-shot translation, where the model translates directions never seen in training, 
the inference-time behavior is often unstable~\cite{gu-etal-2019-improved,al-shedivat-parikh-2019-consistency,rios-etal-2020-subword,raganato-etal-2021-empirical}.
In practice, pivoting through an intermediate language typically gives a strong performance upper bound difficult to surpass by direct zero-shot translation~\cite{al-shedivat-parikh-2019-consistency,arivazhagan2019missing,zhu-etal-2020-language,yang-etal-2021-improving-multilingual}. 
Mapping the source sentences to discrete codes could act as a \textit{pseudo}-pivoting step, which
we hope to make the model more robust under zero-shot conditions.

The discrete codes also provide a new way to interpret model representations.
While there are a wealth of methods to analyze knowledge-sharing in multilingual NMT~\cite{aji-etal-2020-neural,mueller-etal-2020-analysis,chiang-etal-2022-breaking}, 
they mostly either measure translation performance as a proxy,
or
involve sophisticated post-processing after model training, 
e.g. correlation scores between model hidden states~\cite{kudugunta-etal-2019-investigating,chiang-etal-2022-breaking},  training classifiers to probe linguistic features~\cite{liu-etal-2021-improving-zero},
or pruning model submodules~\cite{kim-etal-2021-multilingual}.
In contrast, 
when the model hidden states are directly associated with discrete tokens,
they are directly more \textit{interpretable}.
This characteristic is especially relevant in unseen testing conditions, 
where it is important to pinpoint the underlying cause of model behavior.

\begin{figure}%
\centering
\includegraphics[width=\linewidth]{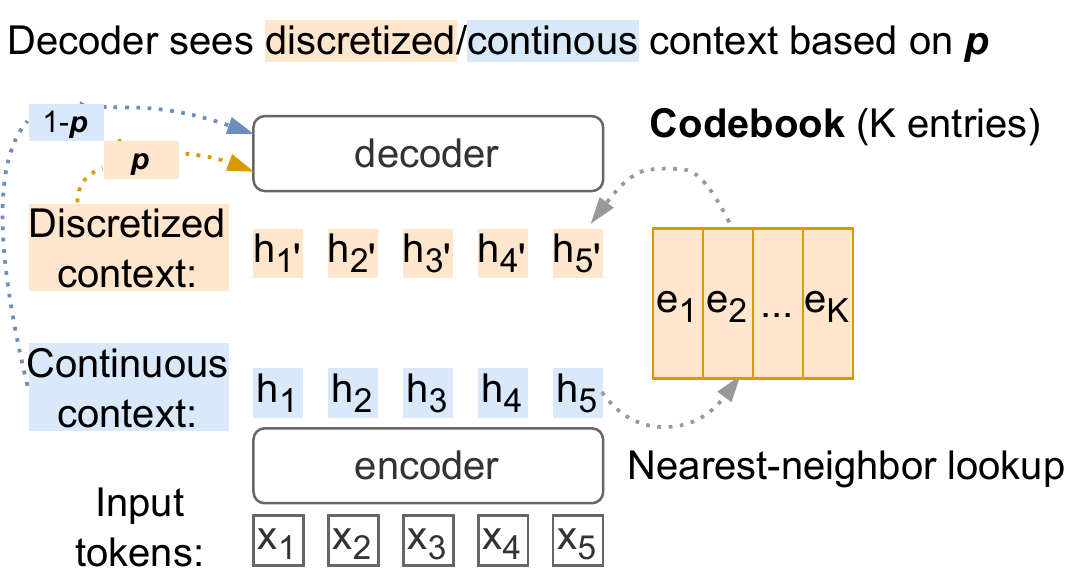}
\caption{
\label{fig:multi} An illustration of our approach, which introduces a codebook for discretizing the encoder output latent space.
During training, the decoder sees discretized and continuous context based on probability $p$.
For inference, we use the continuous context, which have been well-clustered into a set of cluster centers after training.}
\end{figure}

Despite the advantages, 
discretizing  the latent space of NMT models 
makes them inherently less expressive than their fully continuous counterparts.
Maintaining translation performance relative to the continuous models is therefore a challenge.
To strike a balance between expressiveness and discretization, 
we propose a \textit{soft} discretization approach:
In training, we assign each encoder hidden state to an entry in a fixed-size codebook.
This step in effect clusters encoder hidden states to one of the many cluster centers in the latent space.
The codebook where the cluster centers come from
is then trained along with the translation model.
To ensure that the decoder receives sufficient context information,
we make it access both the discretized 
or continuous context, as illustrated in \autoref{fig:multi}.
In our experiments on data from the Large-Scale Multilingual Translation Shared Task~\cite{wenzek-etal-2021-findings} from WMT21~\cite{akhbardeh-etal-2021-findings}, 
our approach is able to
learn meaningful discrete codes and 
achieve translation performance competitive with models with continuous latent spaces.
Our main contributions are:
\begin{itemize}[noitemsep,topsep=0pt]
\item We propose a framework to learn discrete tokens as intermediate representations of multilingual NMT models (\S \ref{sec:approach}).
\item On large-scale multilingual translation experiments, our approach is competitive with strong alternatives while offering more interpretable intermediate representations (\S \ref{subsec:translation-performance}).
\item We use the learned discrete codes to study the role of bridging languages. Using two novel analyses, namely \textit{code overlap} and \textit{code translation}, we discover that using a similar bridge language facilitates knowledge-sharing in all languages covered by the model (\S \ref{subsec:discrete-codes}).
\end{itemize}

\section{Related Work}
\input{sections/related_work.tex}

\section{Learning Discrete Codes} 
\label{sec:approach}
As motivated in \S \ref{sec:intro}, we aim to learn to represent sources sentences with a sequence of discrete codes out of a codebook.
To this end, 
alongisde the translation objective,
we also train our model to partition the continuous latent space of the encoder output into discrete subspaces.
Each of the discrete subspaces is represented by one of the $k$ entries (cluster centers) from a trainable codebook, 
and the encoder hidden states are assigned to these entries.
To learn a meaningful discretization, the learned cluster centers must fulfill some requirements: 
1) avoid trivial solutions where all points are assigned to one or a few codebook entries,
2) carry sufficient context information for the decoder for the translation task, despite being less expressive than the encoder output prior to the discretization step.

\begin{figure}%
\centering
\includegraphics[width=\linewidth]{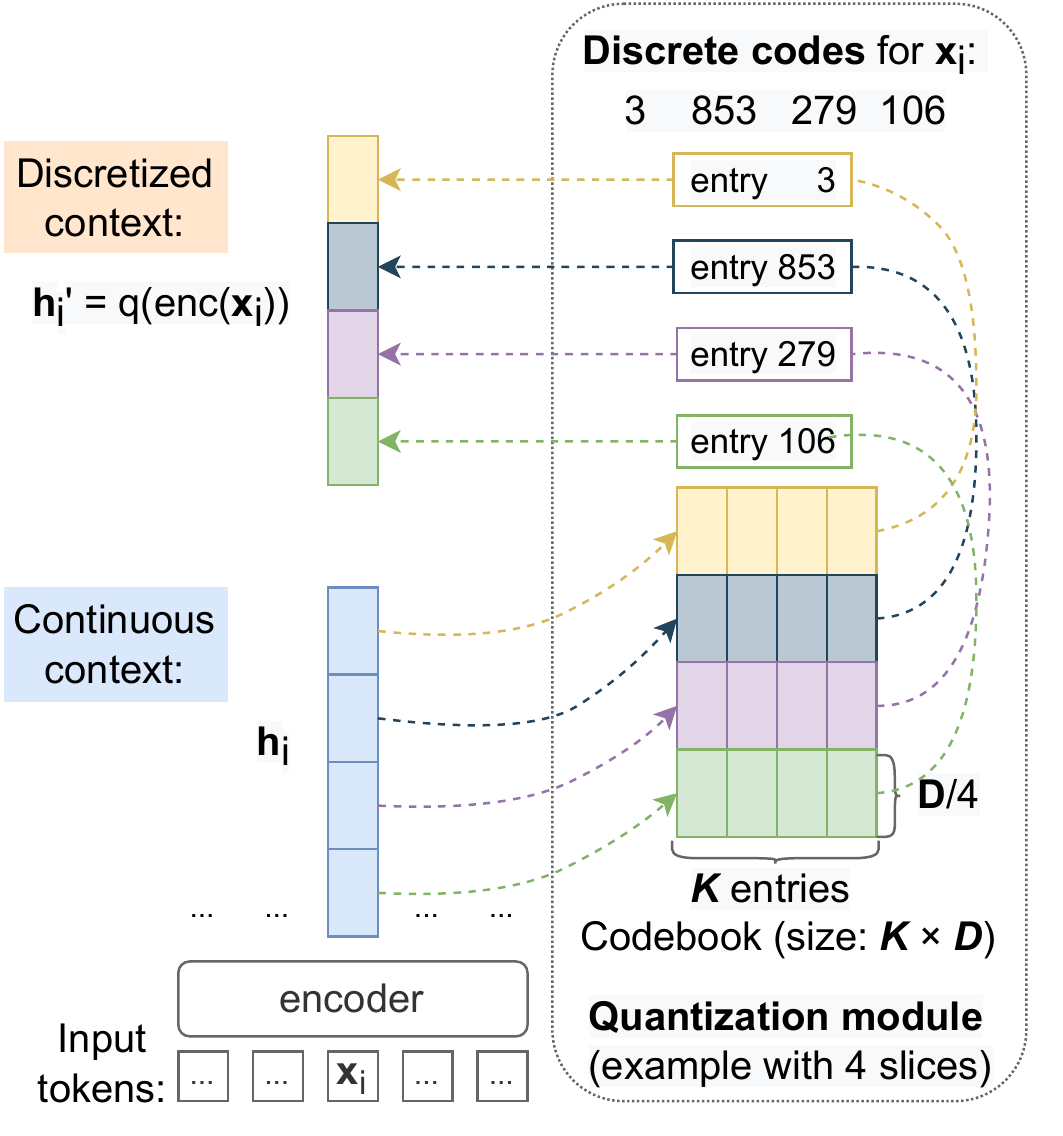}
\caption{
\label{fig:slice} 
Illustration of the generation of the discrete codes based on a sliced \cite{kaiser} codebook.
}
\end{figure}

\subsection{Discretizing Encoder Latent Space}
Compared to a standard Transformer~\cite{transformer}, 
our model includes a quantization module between the encoder and decoder.
We denote the quantization operation as $q(\cdot)$.
Before being passed to the decoder, 
the encoder hidden states $\text{enc}(X)$ for input sequence $X$ first goes through the quantization module, 
which runs a nearest neighbor lookup in an embedding table, i.e. the codebook.
Following the notations from~\citet{vq-vae}, 
the codebook $e \in R^{K \times D}$ has $K$ entries, each with dimensionality $D$.
In our case, $D$ is the same as the embedding dimension of the encoder, resulting in $q(\text{enc}(X))$ with the same shape as $\text{enc}(X)$.

For an input token $X_i$, its quantized representation is one of the $K$ entries from the codebook $e_{k\in [1, K]}$, 
where $k$ is determined by a nearest neighbor search in the embedding space, using the encoder output $\text{enc}(X_i)$ as query:
\begin{equation} 
\label{eq:k}
k = \argmin_{j \in [k]} \| \text{enc}(X_i) - e_j \|_{2},
\end{equation}
where $\|\cdot\|_2$ indicates the Euclidean distance.

The quantization step above is vulnerable to index collapse~\cite{kaiser}, 
where only few entries from the embedding table are actively used.
On auto-encoding tasks,~\citet{kaiser} proposed a countermeasure by breaking down the hidden dimension into multiple slices and quantizing each of them.
Specifically, for input token $X_i$, its encoder hidden state $\text{enc}(X_i)$ is split into $S$ slices: 
\begin{equation}
\text{enc}(X_i)_{1} \oplus \text{enc}(X_i)_{2} \cdots \oplus \text{enc}(X_i)_{S},
\end{equation}
where each slice $\text{enc}(X_i)_{j\in[S]}$ is of $D/S$ dimensions.
A nearest neighbor search is conducted for each slice on the corresponding dimensions in the embedding table.
The results are then concatenated and form the quantized representation:
\begin{equation}
q(\text{enc}(X_i)_{1}) \oplus q(\text{enc}(X_i)_{2}) \cdots \oplus q(\text{enc}(X_i)_{S}),
\end{equation}
and passed to the decoder as context.
\autoref{fig:slice} illustrates this process.

The slicing mechanism resembles multi-head attention~\cite{transformer} in that both split the embedding dimension into subspaces for richer representation. 
Therefore, we will use the same number of slices as the number of attention heads.

\subsection{Soft Discrete Codes}
\paragraph{Training} Compared to encoder outputs in a continuous space, 
the quantization module is an \textit{information bottleneck}.
In practice, limiting the amount of context information passed to the decoder will likely degrade translation quality.
To strike a balance between discretization and performance,
we make the discrete codes \textit{soft}, 
in that the decoder can still access to the richer information prior to quantization by a probability.
Specifically, during training, the encoder gives the quantized context $q(\text{enc}(X))$ by probability $p$, and the raw context $\text{enc}(X)$ by probability $1-p$.
This procedure is illustrated in \autoref{fig:multi}.

In Equation \ref{eq:k}, the lookup of index $k$ is a non-differentiable operation. 
When the encoder passes on the quantized context,
in order to train the parameters below the quantization module, 
we use the straight-through estimator~\cite{straightthrough} to copy gradients onto the pre-quantization encoder outputs.
For the copied gradients to be useful for training,  
the difference between $\text{enc}(X_i)$ and $q(\text{enc}(X_i))$ should be limited. 
To achieve this, we use the codebook loss and commitment loss from VQ-VAE~\cite{vq-vae}:
\begin{equation}
\label{eq:codebook}
\mathcal{L}_{\text{codebook}} = \|\text{sg}[\text{enc}(X)] - q(\text{enc}(X))\|_2
\end{equation}
and 
\begin{equation}
\label{eq:commitment}
\mathcal{L}_{\text{commitment}} = \| \text{enc}(X) - \text{sg}[q(\text{enc}(X))]\|_2,
\end{equation}
where $\text{sg}[\cdot]$ denotes the stop gradient operation.
Intuitively, 
Equation \ref{eq:codebook} pushes the codebook entries closer to the points assigned to them,
while Equation \ref{eq:commitment} limits the growth of the encoder hidden states by clipping them to the codebook entries.
Each of the terms has weights $\alpha_\text{{codebook}}$ and $\alpha_\text{{commitment}}$ to control their importance relative to the main translation objective.

\paragraph{Inference}
After training with this mechanism, 
one can expect that
the encoder hidden states are well-clustered around a set of codebook entries.
At test time, we use the continuous context $\text{enc}(X)$ which still carries more information than the cluster centers represented by the codebook entries.
We will verify this property in later experiments (\S \ref{sec:analyze-codes}).


\section{Experimental Setup}
\label{sec:setup}
To experiment on realistic data volumes, 
we use the parallel data\footnote{\url{https://data.statmt.org/wmt21/multilingual-task/small_task2_filt_v2.tar.gz}} from the Large-Scale Multilingual Machine Translation Shared Task~\cite{wenzek-etal-2021-findings} from WMT 2021~\cite{akhbardeh-etal-2021-findings}.
We focus on small-task-2 on Southeast Asian languages.
To study model robustness in zero-shot conditions and 
the role of language relatedness,
we select parallel data between the two high-resource languages: Indonesian (id) and English (en) and three other languages in the Austronesian family: Javanese (jv), Malay (ms), and Filipino/Tagalog (tl).
This leads to two data conditions:
\begin{itemize}[noitemsep]
\item Indonesian-bridge (\textbf{\textsc{Id-Bridge}})
\item English-bridge (\textbf{\textsc{En-Bridge}})
\end{itemize}

As pretrained initialization has been shown beneficial in many submissions last year~\cite{yang-etal-2021-multilingual-machine,liao-etal-2021-back,xie-etal-2021-tentrans},
we initialize the models with the pretrained \texttt{M2M-124} model provided in the shared task~\cite{wenzek-etal-2021-findings}.
It is worth noting that \texttt{M2M-124} has seen parallel data for our \textit{zero-shot} directions, 
hence zero-shot only describes the condition in our \textit{finetuning} step.
This setup is motivated by the observation that existing pretrained models are often trained on massive amounts of data, which are not always feasible to access or store. 
We therefore treat the pretrained \texttt{M2M-124} as a given resource, without relying on all its training parallel data.
We use this setup to especially study if the models can retain the pretrained knowledge on directions that are zero-shot in finetuning.
\input{tables/dataset_stats.tex}

\input{tables/main.tex}

\subsection{Data} \label{subsec:data}
The training parallel data~\cite{wenzek-etal-2021-findings} are compiled from the OPUS platform~\cite{tiedemann-2012-parallel}.
The specific datasets are listed in \autoref{sec:dataset-details}.
As parts of the training data are crawled and therefore rather noisy, 
we follow the filtering steps opened sourced by~\citet{mm100}, 
including length filtering, bitext de-duplication, and histogram filtering.
An overview of the training data after filtering is in \autoref{tab:stat}.
Following the evaluation protocol of the shared task~\cite{wenzek-etal-2021-findings}, 
we report spBLEU on the FLoRes-101~\cite{10.1162/tacl_a_00474} devtest set.
We additionally report chrF++~\cite{popovic-2017-chrf} as another metric.

\subsection{Baselines}
\label{subsec:baselines}
Besides comparing to directly training on our baseline model,
we also compare to two existing approaches that encourage language-independent representations, both of which have been shown effective in zero-shot translation:

\noindent
\textbf{Language-Independent Objective}~\cite{pham-etal-2019-improving,arivazhagan2019missing}
applies an additional loss function that enforces the representations for the source and target sentences to be similar.
The loss function minimizes the difference between encoded source and target sentences after pooling.
Details about the implementation are in Appendix \ref{sec:appendix-sim-reg}.

\noindent
\textbf{Adversarial Language Classifier}~\cite{arivazhagan2019missing} aims to remove source language signals from the encoder hidden states, and thereby create more language-independent representations.
A language classifier is trained on top of the encoder, 
and its classification performance is used adversarially on the encoder through a gradient reversal layer~\cite{ganin2016domain}.
Details about the implementation are in Appendix \ref{sec:appendix-adv}.


\subsection{Training and Inference Details}
As motivated in \S\ref{sec:setup}, 
we finetune from the small variant of \texttt{M2M-124} with $175$M parameters. 
This model has a vocabulary size of $256$K, $6$ layers in both the encoder and decoder, $16$ attention heads, embedding dimension of $512$ and inner dimension of $2048$.
As the training data for different languages are very unbalanced, we use temperature-based sampling~\cite{arivazhagan2019massively} with coefficient $5.0$, which heavily upsamples low-resource directions and is recommended for unbalanced data conditions~\cite{arivazhagan2019massively,tang-etal-2021-multilingual}.
Additional details are in \autoref{sec:training-details}.

For our codebook approach, we use $10$K codebook entries.
Initial trials with a size of $1$K gave worse performance, while $40$K heavily reduced training speed.
We choose $16$ slices\footnote{Initial experiments on smaller datasets showed weaker translation performance with 2 and 4 slices.} for the codebook, the same value as the number of attention heads. 
We keep these two values identical as both slicing and multi-head attention breaks the embedding dimension into multiple subspaces of lower dimensionality.
The scale on the codebook loss and commitment loss ($\alpha_\text{codebook}$ and $\alpha_\text{commitment}$) are $1.0$ and $1.001$.
We found the model sensitive to increasing $\alpha_\text{commitment}$, 
where higher values leads to index collapse\footnote{A potential reason is the encoder parameters are updated too aggressively by the commitment loss in these cases.}.
After exponentially decreasing it to approach $1.0$, we settled at $1.001$.
For the probability of seeing the continuous encoder context, 
with a search among \{$0.1$, $0.5$, $0.7$ $0.9$\}, 
we found $0.9$ and $0.5$ the best parameters for \textsc{Id-Bridge} and \textsc{En-Bridge} respectively.

We implement our approach and the two baselines (\S \ref{subsec:baselines}) with \textsc{fairseq} \cite{ott-etal-2019-fairseq}.

\section{Main Results}
We first discuss the translation performance of our multilingual systems (\S \ref{subsec:translation-performance}), and then 
use the learned discrete codes to investigate cross-lingual knowledge-sharing of the trained models (\S \ref{subsec:discrete-codes}).

\subsection{Translation Performance} 
\label{subsec:translation-performance}

\paragraph{Baseline Conditions}
To set the upcoming results in context, we first present the performance of training without additional improvements in rows $(1)$-$(3)$ and $(4)$-$(6)$ of \autoref{tab:pretrain_wmt}.
Rows $(1)$ and $(4)$ show the performance of training with random initialization.
This corresponds to a condition where we have parallel data but no pretrained resources.
On the other side of the spectrum, in row $(2)$ and $(5)$, 
we report the results of directly running inference on the pretrained \texttt{M2M-124} model.
This corresponds to another extreme where we have access to pretrained models but cannot additionally train on parallel data.
In rows $(3)$ and $(6)$, we combine the best of two worlds: initializing with pretrained model and finetuning on parallel data.
For supervised directions, pretraining mainly improves $\rightarrow$English directions:
In the \textsc{En-Bridge} condition, initializing with \texttt{M2M-124} gains $1.1$ spBLEU over random initialization, from $27.0$ to $28.1$ spBLEU.
For other supervised directions, however, we do not observe gains from pretraining.
This could be related to the pretrained model being particularly strong at decoding English.
For zero-shot directions in our setup (these directions are seen in training by the pretrained model),
as they are comparatively low-resource among all the directions covered in \texttt{M2M-124}, 
out-of-box translation quality on these directions is relatively low, 
with an average of $9.9$ spBLEU.
However, 
when finetuning,
we see a striking difference between \textsc{Id-Bridge} and \textsc{En-Bridge}: 
there is a large gain from $9.9$ to $17.7$ spBLEU with the former,
but a degradation from $9.9$ to $5.1$ spBLEU for the latter.
We study this phenomenon next.

\paragraph{Impact of Bridge Languages}
For \textsc{En-Bridge}, the finetuning step causes catastrophic forgetting of the zero-shot directions ($-4.8$ spBLEU).
On the other hand, for the \textsc{Id-Bridge} condition, pure finetuning leads to substantial improvements in \textit{both} supervised and zero-shot directions.
The gain from $9.9$ to $17.7$ spBLEU in the $Y{\leftrightarrow}Z$ directions is particularly noteworthy since the model has not seen parallel data for these directions in finetuning.
This indicates that the growth in supervised directions brings zero-shot directions forward too.
Moreover, on these directions, pretraining also gives large gain of $1.9$ spBLEU over random initialization.
Overall, the observations suggest that incorporating a similar language as bridge is beneficial to re-using pretrained knowledge.
Furthermore, given that the amount of parallel data in the \textsc{En-Bridge} condition is nearly $4$ times of that in the \textsc{Id-Bridge} condition,
using a similar bridge language also appears to be more \textit{data-efficient}.
This likely related to all translation directions being similar, therefore easing the multilingual learning task.

\paragraph{Impact of Using Codebooks}
Compared to pure finetuning in rows $(3)$ and $(6)$, 
by incorporating the codebook we improve zero-shot translation by $0.6$ and $10.1$ spBLEU for \textsc{Id-Bridge}  and\textsc{En-Bridge} respectively.
Compared to the two existing approaches, 
namely language-independent objective and adversarial language classifier in rows $(*.1)$ and $(*.2)$, 
our approach performs on par with them for \textsc{Id-Bridge}, 
achieving $18.3$ spBLEU for $Y{\leftrightarrow}Z$ directions and $21.9$ spBLEU over all directions.
In the more challenging \textsc{En-Bridge} condition, 
we fall behind the two other approaches by around $2.0$ spBLEU on zero-shot directions.
Using a language identifier\footnote{\url{https://github.com/facebookresearch/fairseq/tree/nllb##lid-model}} \cite{nllb}, we found that the culprit here is still off-target translation, where some test sentences were translated to an incorrect language.
While our codebook approach reduces the proportion of off-target sentences from $87.4$\% to $13.1$\% compared to the pure finetuning baseline in row $(6)$, 
the figure is still higher than the $4.7$\% achieved by the two alternative models in rows $(6.1)$ and $(6.2)$.
Despite this gap, an advantage of our approach is easier analyses of learned representations, 
which we will now leverage to investigate why the two data conditions come with very distinct zero-shot behavior.

\subsection{Using Discrete Codes to Interpret Learned Representations} \label{subsec:discrete-codes}
Since our codebook approach allows easier interpretation of model hidden representations, we take advantage of this characteristic to answer the following question: 
\textit{why is the \textsc{Id-Bridge} data condition more performant despite using less data?}

\paragraph{Formalization}
To this end, we first extract the discrete codes for all source languages on the test set\footnote{The FLoRes-101 test set is multiway. Therefore the semantic meanings of the sentences are the same.}.
Given a total of $S$ slices,
a sentence with $t$ tokens $X_{1,...,t}$ is represented as $S$ sets of discrete tokens $T_{1,...,t}^{s}$ for slice $s$, where $s{\in}[S]$.
Between two sets of semantically identical sentences (e.g. multiway test sets in two different languages),
we can compare the discrete codes by examining: 1) their overlap and 2) the difficulty of transforming one set to another.
The results quantify the similarity between the two sets of codes, 
and hence the model representations for the two source languages.

\paragraph{Discrete Code Distribution}
For each slice, we normalize the code occurrences into a probability distribution.
The distribution $P$ is defined by:
\begin{equation}
p(c_i)=\frac{\text{frequency}(c_i)}
{\sum_{c_j{\in}[C]} \text{frequency}(c_j) },
\end{equation}
where $c_i$ is a discrete code from the set  $[C]$.
For a pair of languages $i$ and $j$, we then compute the KL divergence between their code distributions $P_i$ and $P_j$:
\begin{equation}
D_{\text{KL}}^{(i,j)}=(P_i || P_j).
\end{equation}



\begin{figure}
    \centering
    \includegraphics[width=\linewidth,keepaspectratio]{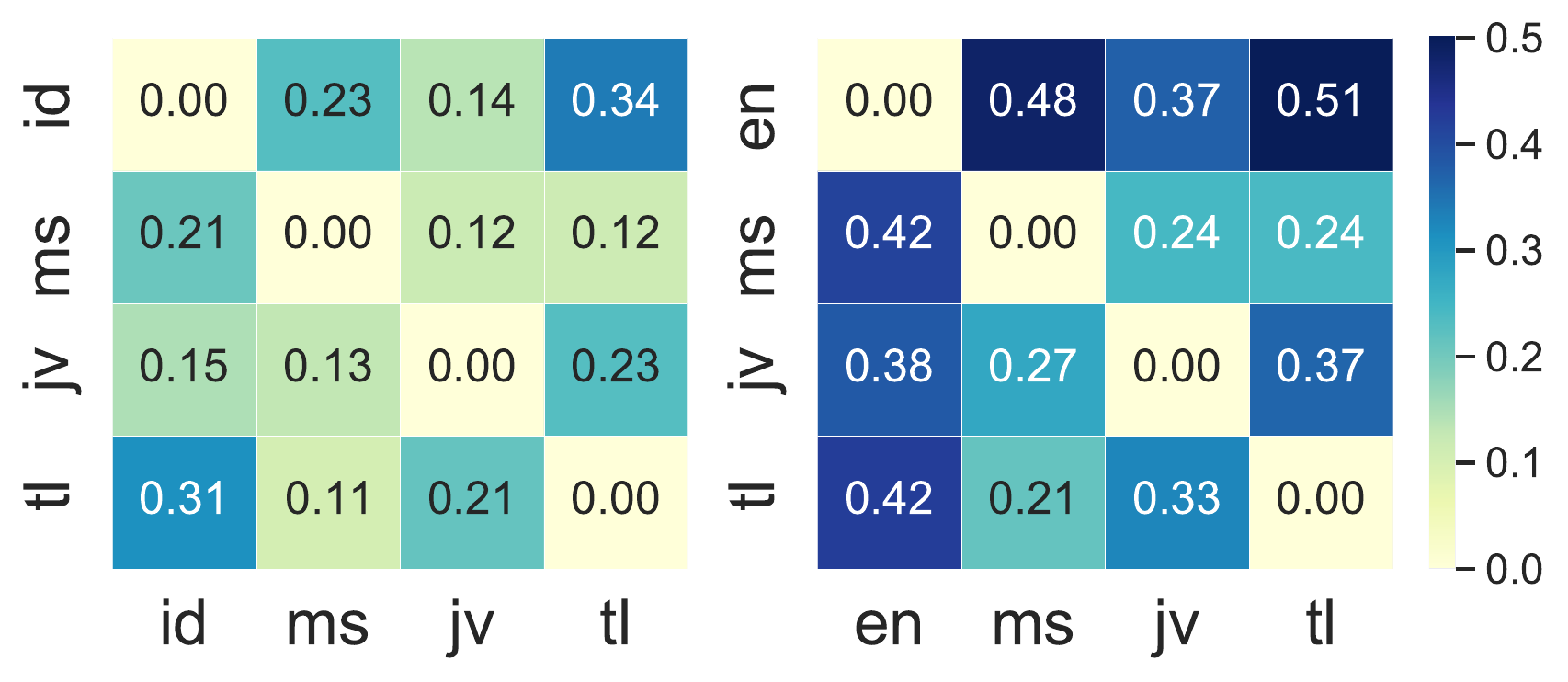}
    \caption{
    KL divergence$(\downarrow)$ of code distribution for the \textsc{Id-Bridge} (left) and \textsc{En-Bridge} (right) setup.
    \textit{Lower} values indicate a higher degree of sharing.
    \textsc{Id-Bridge} results in more sharing not only between itself and \{ms, jv, tl\} but also among \{ms, jv, tl\}. 
    }
    \label{fig:code-overlap}
\end{figure}

\autoref{fig:code-overlap} depicts the KL divergence of code distribution averaged over all slices.
A comparison of the En- and \textsc{Id-Bridge} setup exhibits several major differences.
First, the clearly prominent first row and column in \textsc{En-Bridge} shows that its bridge-language is represented very differently from all other languages (\{ms, jv, tl\}).
For the \textsc{Id-Bridge} counterpart, the difference between the bridge language and the remaining languages is much milder.
Second, but perhaps more importantly, among the languages used in zero-shot directions (\{ms, jv, tl\}), the amount of sharing is also higher under the \textsc{Id-Bridge} setup.
This finding is crucial as the raw tokens for \{ms, jv, tl\} are identical between the \textsc{Id-Bridge} and \textsc{En-Bridge} setup.
Therefore, the higher degree of sharing is clearly an outcome of the model creating its representations differently.
Overall, these results show that 
the choice of the bridge language not only impacts the knowledge-sharing mechanism between itself and the remaining languages, but also for the remaining languages in the model.


\paragraph{Discrete Code Translation}
The code distribution analysis above makes a simplified assumption by considering the discrete codes as a \textit{bag of words}.
To additionally assess the \textit{structural (dis)similarity} between the code representations for different languages,
we consider the task of \textit{translating} the discrete codes of a language to another.

While a constructed language like Interlingua would create the same representations for the source sentences with identical meanings, 
our discrete code representation is not yet invariant to the source language.
Nevertheless, we do expect them to be more abstracted from the source sentences,
making the translation task easier than directly between the raw tokens.
Here we train a translation model on the discrete codes and use the test performance to quantify how similarly the source languages are represented.
When the representations are more different from each other, i.e. language-specific, 
the translation quality on the discrete is expected to be lower.

\begin{figure}
    \centering
    \includegraphics[width=\linewidth,keepaspectratio]{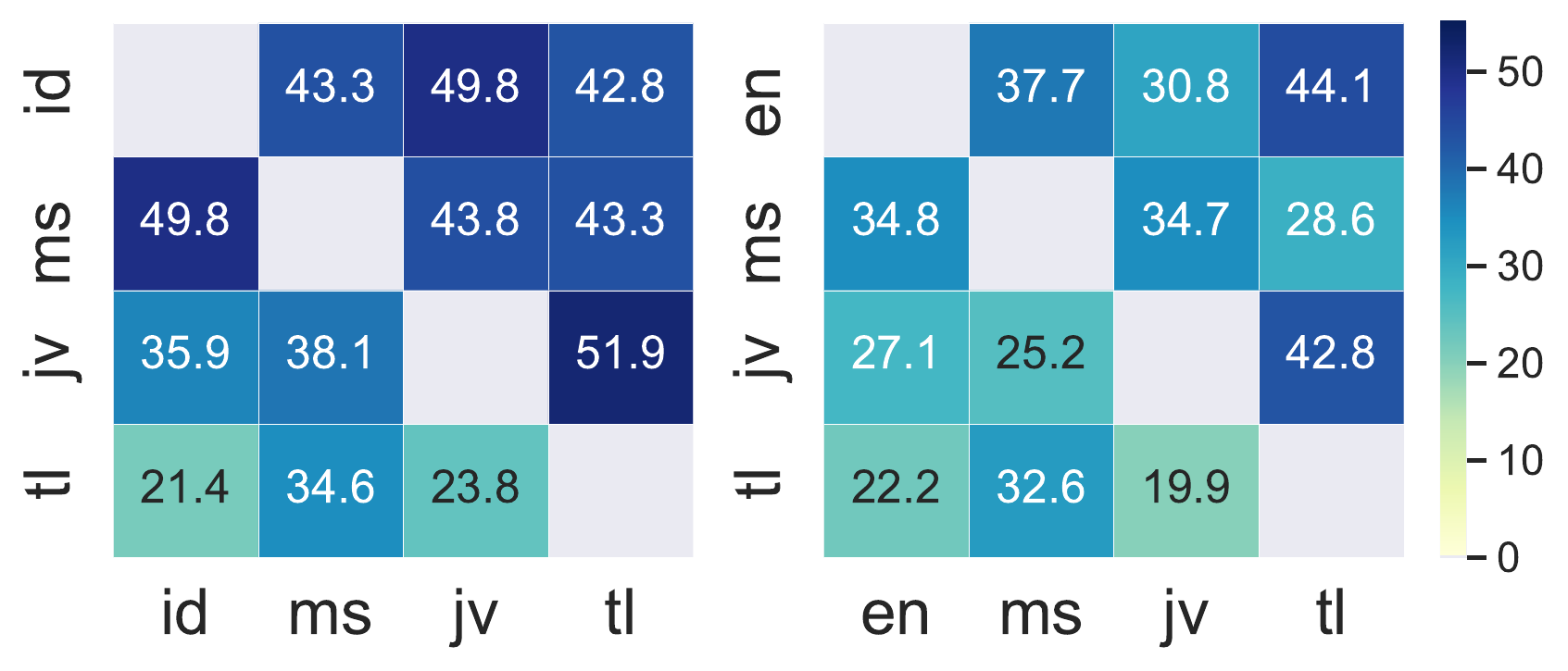}
    \caption{BLEU($\uparrow$) scores of translating between discrete codes for the \textsc{Id-Bridge} (left) and \textsc{En-Bridge} (right) setup.
    \textit{Higher} values indicate a higher degree of sharing.
    In general it is easier to translate the codes for the \textsc{Id-Bridge} setup, 
    indicating more structural similarity between the representations.}
    \label{fig:code-translation}
\end{figure}

Specifically, 
we randomly sample $100$K sentence pairs\footnote{The training data (\autoref{tab:stat}) allow us to use $340$K sentences. We sampled $100$K for faster experiment iteration.} for each translation direction in the experiments of \autoref{tab:pretrain_wmt}
extract their discrete codes assigned by the trained models (rows ($3.3$) and ($6.3$) of \autoref{tab:pretrain_wmt}), 
and train a new Transformer-base~\cite{transformer} to translate between the extracted codes of different languages.
We flatten the slices, therefore making each source token represented by $16$ discrete codes.
After training for $200$K steps, we report BLEU scores on the test set, which is also converted to discrete codes.
The results are shown in \autoref{fig:code-translation}.
First, the translation task is clearly easier on the discrete codes derived from the \textsc{Id-Bridge} system.
Second, the scores differences are especially prominent when translating out of Malay (ms) and Javanese (jv), which are more related to Indonesian than Filipino/Tagalog (tl).
Along with the results from the code overlap, 
our results show that 
using a similar bridge language results in higher knowledge-sharing not only syntactically but also structurally, especially between related languages.

\section{Analyses on Learned Discrete Codes} \label{sec:analyze-codes}
Next we further investigate the discrete codes regarding 
its usefulness for the learned representations (\S \ref{subsec:cluster}) as well as the translation task (\S \ref{subsec:center}).


\subsection{How well-clustered are the hidden states?}
\label{subsec:cluster}
\begin{figure}
    \centering
    \subfloat[\centering \textsc{Id-Bridge}]{{\includegraphics[width=3.25cm]{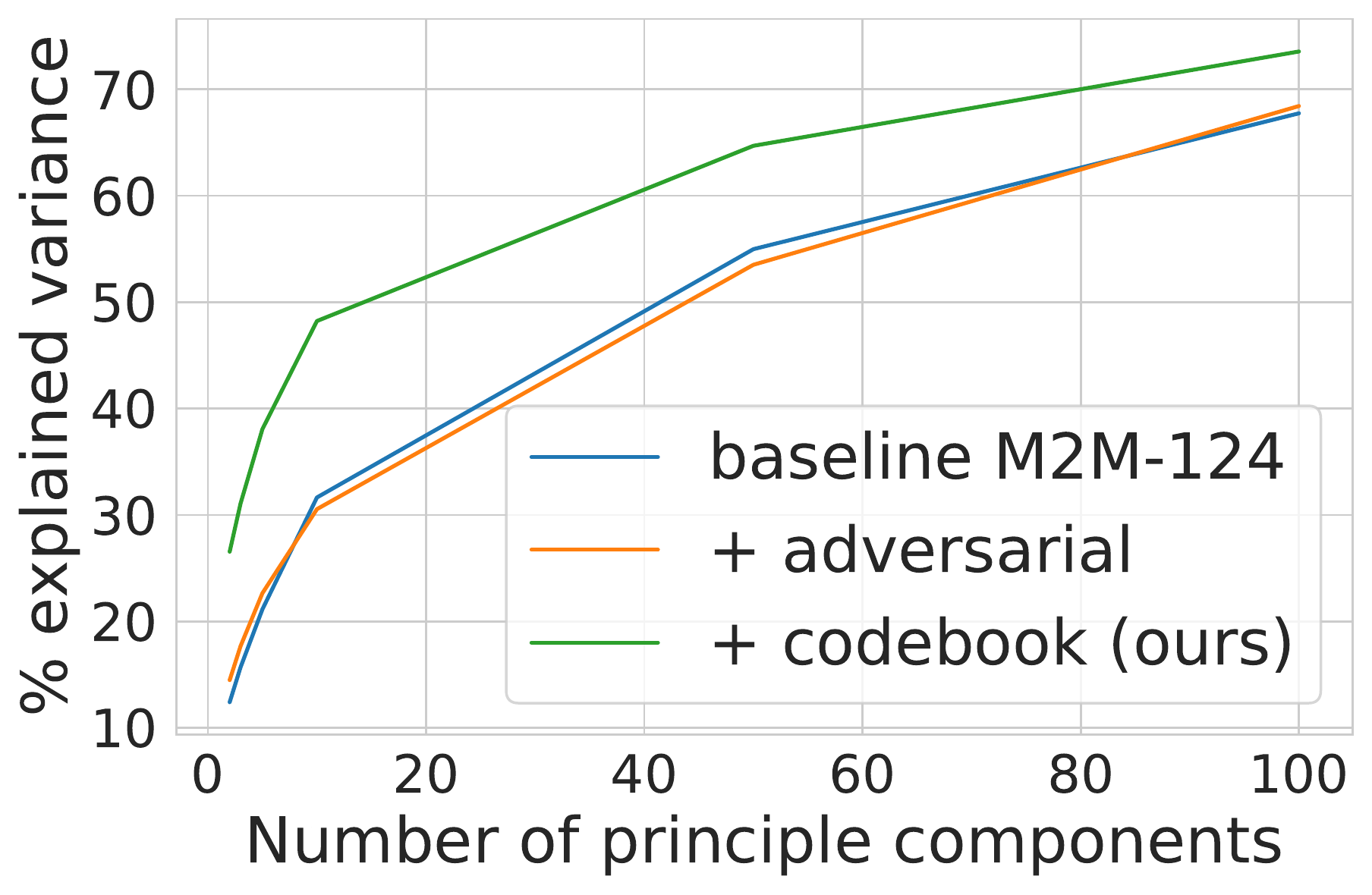} }}%
    \qquad
    \subfloat[\centering \textsc{En-Bridge}]{{\includegraphics[width=3.25cm]{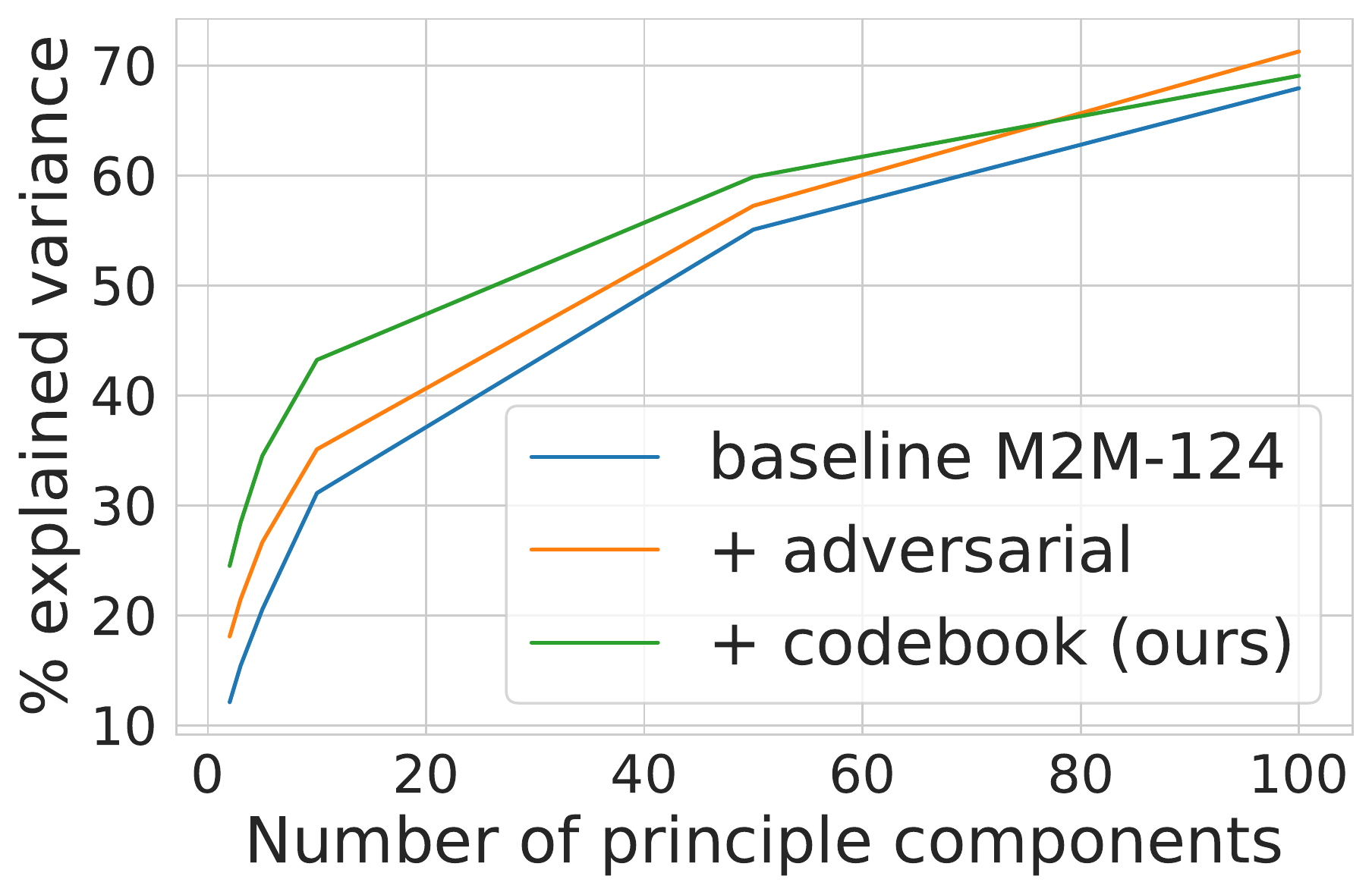} }}%
    \caption{Our codebook approach creates better-clustered encoder hidden states, 
    as shown by a much higher percentage of variance explained by PCA compared to both the baseline and a strong alternative approach (adversarial language classifier).}
    \label{fig:pca}
\end{figure}

As motivated in \S \ref{sec:approach}, 
although at inference time we use the continuous encoder hidden states instead of the cluster centers, 
the soft discrete codes will still enforce encoder hidden states into clusters, 
thereby resembling a discrete structure.
To verify whether the encoder latent space indeed becomes more discretized with our approach,
we analyze the encoder hidden states on the test set using Principle Component Analysis (PCA).
If the data points representing the encoder outputs are well-clustered, 
a larger percentage of their variance should be explained by the learned principle components.
As shown in \autoref{fig:pca},
our approach (marked with green line) consistently leads to higher proportions of explained variances compared to the baseline \texttt{M2M-124}, as well as the strong alternative approach with the adversarial language classifier.
These results therefore confirm the effectiveness of our soft discrete code approach in enforcing discrete structures in the encoder latent space.

\subsection{Meaningfulness of Clusters Centers}
\label{subsec:center}

Recall that at inference time our soft discrete code model uses the encoder hidden states prior to discretiztaion, 
although it does use  both pre- and post-discretization encoder context in training.
A main reason of doing so is that discretizing the encoder hidden states to cluster centers creates an information bottleneck that limits model expressiveness.
Despite the expected performance degradation, we are nonelessness interested in \textit{quantifying} how much information is lost by using the cluster centers as context instead.
In other words, the question is \textit{how meaningful are the cluster centers for the translation task?}
In \autoref{tab:cluster-center}, we report the results of using the cluster centers as context for the decoder at inference time.
Compared to using the encoder hidden states, we see a degradation of  $4.1$ and $1.7$ and spBLEU for \textsc{Id-Bridge} and \textsc{En-Bridge} respectively.
This indicates that the cluster centers are still relevant for the translation task, although much less powerful than the encoder hidden states prior to discretization.
It also rules out the possibility of the learned codes being trivial repetitions, 
which would otherwise have been detrimental to the translation performance.

\input{tables/cluster_center.tex}

\section{Analyses on Zero-Shot Translation}
Our experiments so far use single-bridge languages
and are evaluated in part on zero-shot directions.
We now study the impact when either of the two conditions changes: 
1) when parallel data is available for previously zero-shot directions; 2) when using multiple bridge languages.

\subsection{When does zero-shot translation match the performance on parallel data?}

Zero-shot conditions could be avoided by creating synthetic data from back-translation~\cite{sennrich-etal-2016-improving,zhang-etal-2020-improving} or mining additional parallel data~\cite{mm100, freitag-firat-2020-complete}.
Both approaches introduce additional workflows into the pipeline of building translation systems.
We are therefore interested in the following question: \textit{How much parallel data do we need to perform better than direct zero-shot translation?}

The training corpora from the shared task (\S \ref{subsec:data}) provides an oracle condition to answer this question.
As shown in~\autoref{tab:stat},
the oracle parallel data amounts to $2.2$M sentences in total 
($340$K for jv-ms, $662$K for jv-tl, and $1.2$M sentences for ms-tl).
We take $100$\%, $10$\% and $1$\% of the oracle parallel data 
and training systems together with the original data and train multilingual systems with the same configuration as rows $(3)$ and $(6)$ of \autoref{tab:pretrain_wmt}. 
The results are shown in~\autoref{tab:oracle-zs}.

To our surprise, adding $1$\% oracle bitext ($22$K sentence pairs in total) of the previously zero-shot directions already results in comparable performance to the best zero-shot performance 
($18.4$ and $17.3$ spBLEU for \textsc{Id-Bridge} and \textsc{En-Bridge} respectively).
However, this comes with some degradation on supervised directions of $0.4$ spBLEU for \textsc{Id-Bridge} and $0.7$ for \textsc{En-Bridge}.
This is likely due to the temperature-based sampling aggressively upsampling the extremely low-resource directions, 
meanwhile causing the model to de-prioritize other higher-resource directions.
When increasing oracle bitext to 10\% ($220$K sentence pairs in total), the system outperforms direct zero-shot performance.
Lastly, the additional gain appear to diminish when going from 10\% to all oracle data.
For \textsc{Id-Bridge}, the performance appears saturated at 10\%: adding the remaining 90\% parallel data does not give additional gain.
On the contrary, 
For \textsc{En-Bridge}, the system appears to still improve, especially on $Y{\leftrightarrow}Z$ directions ($+0.5$ spBLEU).
The performance on these directions nevertheless still falls behind the \textsc{Id-Bridge} direction by $0.8$ spBLEU ($18.9$ vs $19.7$ spBLEU).
An explanation is that the \textsc{En-Bridge} system requires more data to train
as a result of the bridge language being very distant to the rest, 
thereby increasing the difficulty of multitasking over all the translation directions.
This echos with the previous finding that using related bridge languages eases the multilingual translation task and increases knowledge-sharing (\S \ref{subsec:translation-performance}).

\input{tables/oracle_zs.tex}

\input{tables/multi_bridge.tex}

\subsection{Do multiple bridge languages bring additional gains?}

While the experiments so far are based on single bridge languages, 
in practice we often have access to multi-bridge parallel data.
Indeed, recent works~\cite{freitag-firat-2020-complete,mm100} have shown success on large-scale fully-connected models, 
as well as evidence of multi-bridge outperforming the English-bridge condition~\cite{rios-etal-2020-subword}.
What remains unclear is whether there is a synergy when combining the parallel data from several single-bridge conditions.
We investigate this hypothesis by training a multi-bridge system, combing the data from our \textsc{Id-Bridge} and \textsc{En-Bridge} setup.
As shown in \autoref{tab:multibridge},
for supervised directions of $\rightarrow$$X$ and $X$$\rightarrow$,
there is no clear difference between the performance of
the multi-bridge system and that of the single-bridge ones.
For zero-shot directions ($Y{\leftrightarrow}Z$), 
while multi-bridge gains substantially over \textsc{En-Bridge} ($18.3$ from $5.1$ spBLEU), 
there is only a slight gain over \textsc{Id-Bridge}.
Given that the multi-bridge model more than doubles the training time of \textsc{Id-Bridge}, 
the little performance difference to the multi-bridge system shows that 
choosing a bridge language related to the remaining languages
is a data-efficient way to achieve strong zero-shot performance.

\section{Conclusion}
In this work, we focus on learning to represent source sentences of multilingual NMT models by discrete codes.
On multiple large-scale experiments, 
we show that our approach not only increase the model robustness in zero-shot conditions, 
but also offers more interpretable intermediate representations.
We leverage the latter property to investigate the role of bridge languages, 
and show that  using a more related bridge language leads to increased knowledge-sharing, 
not only between the bridge language and remaining but also between all other languages involved in training.

A limitation is that the discrete codes only give a mechanism to compare hidden representations,
but are not directly interpretable by humans.
A potential improvement would be to use an existing codebook that corresponds to an actual human language.
Besides this, as next steps, we plan to improve the generation process of the discrete codes. 
The first direction is to make the code lookup conditionally-dependent along the time dimension and learn to shrink the sequence length of the discrete codes, thereby creating a more compact representation.
Another direction is to explicitly incentivize more shared codes between different, and especially related, languages during training.
This would bring the discrete codes closer to a language-independent representation.

\paragraph{Acknowledgement}
We thank James Cross and Paco Guzm\'{a}n for their feedback in the early stage of this work.
We thank the anonymous reviewers for their detailed and insightful feedback.
We also thank Ngoc Quan Pham, Tu Anh Dinh, and Sai Koneru for feedback on the paper draft.

\bibliography{anthology,custom}
\bibliographystyle{acl_natbib}

\appendix

\section{Additional Training and Inference Details} \label{sec:training-details}
When training, one optimization step happens after $16384$ tokens.
We use the Adam optimizer with betas $(0.9, 0.98)$.
The learning rate is $0.0001$ with the inverse squared root schedule and $2500$ warmup steps.
As for regularization parameters, 
we use label smoothing of $0.1$, dropout of $0.3$, and attention dropout $0.1$.
The models are trained for $500$K updates in total.
An exception is the \textsc{Multi-Bridge} experiment with more training data, where we trained for $800$K updates in total.
For inference, we decode with a beam size of $5$.

\section{Dataset Details} \label{sec:dataset-details}
The training parallel data include the following corpora: 
bible-uedin~\cite{bible-uedin}, 
(Multi)CCAligned~\cite{el-kishky-etal-2020-ccaligned}, 
Gnome\footnote{\url{https://opus.nlpl.eu/GNOME.php}}, 
ELRC\footnote{\url{https://opus.nlpl.eu/ELRC.php}}, 
KDE4\footnote{\url{https://opus.nlpl.eu/KDE4.php}}, 
GlobalVoices\footnote{\url{https://opus.nlpl.eu/GlobalVoices.php}}, 
OpenSubtitles\footnote{\url{https://opus.nlpl.eu/OpenSubtitles-v2018.php}}, 
QED~\cite{abdelali-etal-2014-amara}, 
MultiParaCrawl\footnote{\url{https://opus.nlpl.eu/MultiParaCrawl.php}}, 
TED2020\footnote{\url{https://opus.nlpl.eu/TED2020.php}}, 
Tanzil\footnote{\url{https://opus.nlpl.eu/Tanzil.php}}, 
Tatoeba\footnote{\url{https://opus.nlpl.eu/Tatoeba.php}}, 
Ubuntu\footnote{\url{https://opus.nlpl.eu/Ubuntu.php}}, 
WikiMatrix~\cite{schwenk-etal-2021-wikimatrix}, 
wikimedia\footnote{\url{https://opus.nlpl.eu/wikimedia.php}}, 
and TICO-19~\cite{anastasopoulos-etal-2020-tico}.

\section{Implementation of Baselines}

\subsection{Language-Independent Objective} \label{sec:appendix-sim-reg}
We chose meanpool and L2 distance for the similarity loss since it gave better or more consistent performance in initial experiments.
As for the weight of the language-independent objective, 
we used $0.1$\footnote{The $1.0$ in the conference proceeding was a typo.} following \citet{pham-etal-2019-improving}.

\subsection{Adversarial Classifier} \label{sec:appendix-adv}

We extend the adversarial language classification approach from~\citet{arivazhagan2019missing} for robust training.
Specifically, we use a modified loss when adversarially training the encoder.
Moreover, we apply the language classification on the token level to remove the need for selecting a pooling method.
The classifier minimizes the cross-entropy loss when predicting the language labels: 
\begin{equation} \label{eq:loss1}
\mathcal{L}_\text{classifier} = -\sum_{c=1}^{L} y_c \text{log}(p_c),
\end{equation}
where $L$ is the number of classes to predict, 
$y_c$ is a binary indicator whether the true language label is $c$, and $p_c$ is the predicted probability for the instance belonging to language $c$.

Removing source language signals from the encoder representations can be achieved by a gradient reversal layer~\cite{ganin2016domain} from the language classification.
An issue with the standard classification loss in \autoref{eq:loss1} is that, 
when the classifier is performing well, the loss landscape is rather flat, causing minimal gradient flow to the encoder.
In fact, when the classifier predicts the source languages accurately, we instead need large gradients to update the encoder representations as they contain high amounts of language signals.
Therefore, when updating the encoder parameters adversarially, we use the modified loss:
\begin{equation} \label{eq:loss2}
\mathcal{L}_\text{adv\_classifier} = \sum_{c=1}^{L} y_c \text{log}(1 - p_c),
\end{equation}
which in effect mirrors \autoref{eq:loss1} by the horizontal axis and the vertical line defined by $x=0.5$.
With the modified loss, the optimization direction does not change, but the gradient is larger when the classifier is performing well.

The translation model is then trained with:
\begin{equation}\label{eq:loss3}
\mathcal{L}_\text{encoder\_decoder} = \mathcal{L}_\text{MT} + \mathcal{L}_\text{adv\_classifier}.
\end{equation}

For training stability, we alternate the optimization of the classifier (\autoref{eq:loss1}) and the main encoder-decoder parameters (\autoref{eq:loss3}).
Optimizing them jointly would otherwise lead to co-adaptation of the parameters of the translation and classification module
and empirically causes training instability.

\end{document}

%% file: sections/related_work.tex
\paragraph{Multilingual Machine Translation}
Multilingual translation models are able to multitask over many language pairs.
For this large-scale multi-task learning problem, training data plays a critical role.
Low-resource directions often need upsampling to perform well \cite{arivazhagan2019massively,tang-etal-2021-multilingual},
which, meanwhile, brings capacity bottlenecks \cite{aharoni-etal-2019-massively} to high-resource languages.
This capacity bottleneck can be eliminated by dedicated language-specific capacity \cite{bapna-firat-2019-simple,philip-etal-2020-monolingual,moe,share-or-not}.
When scaling up translation coverage \cite{aharoni-etal-2019-massively,zhang-etal-2020-improving,mm100}, zero-shot directions that have not seen any parallel training data is more likely to get encountered.
While many dedicated models or objectives have been proposed to improve the zero-shot performance \cite{al-shedivat-parikh-2019-consistency,arivazhagan2019missing,pham-etal-2019-improving,zhu-etal-2020-language,son-lyu-2020-sparse,liu-etal-2021-improving-zero,yang-etal-2021-improving-multilingual,raganato-etal-2021-empirical}, there is in general a \textit{tradeoff} between supervised and zero-shot performance.

\paragraph{Robustness in Zero-Shot Conditions}
Zero-shot generalization is a widely-discussed direction in machine learning research~\cite{zs_ml,zs_ml_4,zs_ml_3,zs_ml_2}.
In the context of NMT, 
early multilingual models already possess some capability of zero-shot translation of directions unseen in training~\cite{ha-etal-2016-toward,johnson-etal-2017-googles}.
However, 
zero-shot performance has been shown highly sensitive to, among other factors, training data diversity \cite{rios-etal-2020-subword}, 
language token strategies \cite{wu-etal-2021-language,elnokrashy2022language}, and
dropout configurations \cite{arivazhagan2019missing, liu-etal-2021-improving}.
A main cause of the degraded quality is that the zero-shot inference generates \textit{off-target} translation \cite{zhang-etal-2020-improving} into a language other than the desired one.
In recent shared tasks \cite{anastasopoulos-etal-2021-findings,libovicky-fraser-2021-findings}, 
generating synthetic data by back-translation \cite{sennrich-etal-2016-improving} to eliminate zero-shot conditions has been a dominant approach for improving upon pure unsupervised settings \cite{pham-etal-2021-multilingual,zhang-sennrich-2021-edinburghs,liu-niehues-2021-maastricht,knowles-larkin-2021-nrc,libovicky-fraser-2021-lmu}.
A main motivating factor for converting zero-shot conditions to semi-supervised ones is that the latter provides more robust and consistent inference-time behavior.
In this light, to fully realize the potential of knowledge-sharing in multilingual NMT, 
improving zero-shot robustness is an essential task.

\paragraph{Discrete Representations}
Vector Quantized Variational Autoencoder (VQ-VAE;~\citealt{vq-vae}) learns discrete tokens for continuous inputs such as images and audio, and showed its effectiveness in creating discrete representations for speech representations on practical tasks~\cite{DBLP:conf/interspeech/TjandraS020,DBLP:conf/nips/BaevskiZMA20}.
\citet{kaiser} proposed an improvement to VQ-VAE by \textit{slicing}, i.e. decomposing to quantization input and output into several subspaces.
The sliced variant was used in auto-encoding for learning shorter sequences, which allows to accelerate the target generation in auto-regressive decoders.
The most related work to ours is probably that of~\citet{https://doi.org/10.1002/asi.24395},
who used sliced VQ-VAE \cite{kaiser} on translation tasks.
The main difference is that our focus is fully parameter shared multilingual systems while~\citet{https://doi.org/10.1002/asi.24395} focused on auto-encoding and bilingual systems using language-specific encoders and decoders. 
Therefore, in~\citet{https://doi.org/10.1002/asi.24395} zero-shot translation only occurs after a subsequent training step on dedicated encoder for the new language.
Moreover, our approach extends sliced VQ-VAE \cite{kaiser} by soft codes that utilizes both continuous and quantized encoder hidden states.

%% file: tables/dataset_stats.tex
\begin{table}
\centering
\small
\setlength\tabcolsep{7pt} 
\begin{tabular}{lrrrrrr}
\toprule
& \textbf{jv} & \textbf{ms} & \textbf{tl} & \textbf{id} &  \textbf{en}\\ 
\textbf{jv} 
& \cellcolor[HTML]{d3d1d1}
& 340K  
& 662K
& \cellcolor[HTML]{f5f3f3}{644K}
& \cellcolor[HTML]{f5f3f3}{2,556K} \\
\textbf{ms} 
& 2M
& \cellcolor[HTML]{d3d1d1}
& 1,174K 
& \cellcolor[HTML]{f5f3f3}{4,060K}
& \cellcolor[HTML]{f5f3f3}{12,023K} \\
\textbf{tl} 
& 3M
& 16M
& \cellcolor[HTML]{d3d1d1}
& \cellcolor[HTML]{f5f3f3}{2,356K}
& \cellcolor[HTML]{f5f3f3}{12,348K} \\
\textbf{en}
& \cellcolor[HTML]{f5f3f3}18M
& \cellcolor[HTML]{f5f3f3}230M
& \cellcolor[HTML]{f5f3f3}158M
& \cellcolor[HTML]{d3d1d1}
& \cellcolor[HTML]{d3d1d1}  \\
\textbf{id} 
& \cellcolor[HTML]{f5f3f3}{5M}
& \cellcolor[HTML]{f5f3f3}{65M}
& \cellcolor[HTML]{f5f3f3}{30M}
& \cellcolor[HTML]{d3d1d1}
& \cellcolor[HTML]{d3d1d1}  \\
\bottomrule
\end{tabular}
\caption{\label{tab:stat} Number of sentence pairs (above diagonal) 
and target tokens (below diagonal) from bitext 
for each languages pair after preprocessing.
Data marked with \colorbox{gray!10}{light gray} are used in the main experiments.
} 
\end{table}

%% file: tables/main.tex
\begin{table*}[ht]
\small
\centering
\setlength\tabcolsep{5.5pt} 
\begin{tabular}{ llcccccc } 
\toprule
 \textbf{ID} & 
 \textbf{Model} & 
 \multicolumn{4}{c}{\textbf{Avg. spBLEU$(\uparrow)$ (left) and chrF++$(\uparrow)$ (right)}} \\  
 \cmidrule{3-6} 
 &
 &
 $\{$jv,  ms, tl$\}$ \textbf{$\rightarrow$X} & 
 \textbf{X$\rightarrow$} $\{$jv,  ms, tl$\}$ & 
 \textbf{Y$\leftrightarrow$Z} &
 \textbf{Avg. (all dir.)}\\
 \midrule
 &  \multicolumn{4}{l}{\textbf{\textsc{Id-Bridge}} (\textbf{X$=$id})} \\
 $(1)$ &
 random initialization
                    & 27.5 \hspace{4pt} 52.7
                    & 24.2 \hspace{4pt} 49.4
                    & 15.8 \hspace{4pt} 41.5 
                    & 20.8 \hspace{4pt} 46.3 \\
 $(2)$ &
 \texttt{M2M-124} \cite{mm100, 10.1162/tacl_a_00474} 
                    & 20.0 \hspace{4pt} 45.7
                    & 14.7 \hspace{4pt} 38.9
                    & \hspace{2pt} 9.9  \hspace{4pt} 34.3
                    & 13.6 \hspace{4pt} 38.3 \\
 $(3)$ &
 $\hookrightarrow$ parallel data (no data for \textbf{Y}$\leftrightarrow$\textbf{Z})     
                    & 27.1 \hspace{4pt} 52.5
                    & 24.2 \hspace{4pt} 49.6
                    & 17.7 \hspace{4pt} 43.3 
                    & 21.7 \hspace{4pt} 47.2 \\
 $(3.1)$ & 
 \hspace{8pt} $+$ language-independent objective    
                    & 27.1 \hspace{4pt} 52.4
                    & 24.2 \hspace{4pt} 49.6
                    & 18.4 \hspace{4pt} 43.8
                    & 22.0 \hspace{4pt} 47.4 \\
 $(3.2)$ & 
 \hspace{8pt} $+$ adversarial language classifier    
                    & 27.5 \hspace{4pt} 52.9
                    & 24.1 \hspace{4pt} 49.6
                    & 18.4 \hspace{4pt} 44.2
                    & 22.1 \hspace{4pt} 47.7 \\
 $(3.3)$ &
 \hspace{8pt} $+$ codebook (ours)
                    & 27.2 \hspace{4pt} 52.4   
                    & 23.6 \hspace{4pt} 49.2   
                    & 18.3 \hspace{4pt} 44.0        
                    & 21.9 \hspace{4pt} 47.4 \\
\midrule
& \multicolumn{4}{l}{\textbf{\textsc{En-Bridge}} (\textbf{X$=$en})} \\
 
 $(4)$ &
 random initialization       
                    & 27.0 \hspace{4pt} 51.1
                    & 27.8 \hspace{4pt} 51.6
                    & \hspace{2pt} 6.8 \hspace{4pt} 24.5
                    & 17.1 \hspace{4pt} 37.9 \\
 $(5)$ &
 \texttt{M2M-124} \cite{mm100, 10.1162/tacl_a_00474} 
                    & 19.6 \hspace{4pt} 43.6
                    & 14.0 \hspace{4pt} 37.5 
                    & \hspace{2pt} 9.9  \hspace{4pt} 34.3
                    & 13.3 \hspace{4pt} 37.4 \\
 $(6)$ & 
 $\hookrightarrow$ parallel data (no data for \textbf{Y}$\leftrightarrow$\textbf{Z})
                    & 28.1 \hspace{4pt} 51.8
                    & 27.6 \hspace{4pt} 51.8
                    & \hspace{2pt} 5.1 \hspace{4pt} 20.3
                    & 16.5 \hspace{4pt} 36.1 \\
 $(6.1)$ & 
 \hspace{8pt} $+$ language-independent objective    
                    & 27.9 \hspace{4pt} 51.7
                    & 27.2 \hspace{4pt} 51.4
                    & 17.3 \hspace{4pt} 42.8
                    & 22.4 \hspace{4pt} 47.2 \\
 $(6.2)$ &
 \hspace{8pt} $+$ adversarial language classifier    
                    & 27.6 \hspace{4pt} 51.5
                    & 27.1 \hspace{4pt} 51.5
                    & 17.2 \hspace{4pt} 42.8
                    & 22.3 \hspace{4pt} 47.2 \\
 $(6.3)$ &
 \hspace{8pt} $+$ codebook (ours)
                    & 26.8 \hspace{4pt} 50.6
                    & 26.3 \hspace{4pt} 50.9
                    & 15.2 \hspace{4pt} 39.3
                    & 20.9 \hspace{4pt} 45.0 \\

\bottomrule
\end{tabular}
\caption{
\label{tab:pretrain_wmt} 
Translation quality in spBLEU$(\uparrow)$ and chrF++$(\uparrow)$.  
``$\hookrightarrow$'' indicates finetuning on the parallel data (\textsc{Id-Bridge} or \textsc{En-Bridge}; \S \ref{sec:setup}).
Pivoting through the bridge language for $Y{\leftrightarrow}Z$ directions scores $19.7$, $17.5$ spBLEU and $44.9$, $42.8$ chrF++ 
for \textsc{Id-Bridge} and \textsc{En-Bridge} respectively using the systems in rows $(1)$ and $(4)$.
}
\end{table*}








%% file: tables/cluster_center.tex
\begin{table}[ht]
\small
\centering
\setlength\tabcolsep{3pt} 
\begin{tabular}{ llcccccc } 
\toprule
 \textbf{Encoder States at Inference} & 
 \multicolumn{4}{c}{\textbf{Avg. spBLEU}$(\uparrow)$} \\ 
 \cmidrule{2-5} 
 &
 \textbf{$\rightarrow$X} & 
 \textbf{X$\rightarrow$} & 
 \textbf{Y$\leftrightarrow$Z} &
 \textbf{Avg.}\\
 \midrule
 \multicolumn{4}{l}{\textbf{\textsc{Id-Bridge}} (\textbf{X$=$id})} \\
encoder states (Tab. \ref{tab:pretrain_wmt} row $(3.3)$) &
27.2     & 23.6      & 18.3      & 21.9 \\
cluster centers & 
22.8     & 20.0      & 14.3      & 17.8 \\
\midrule
\multicolumn{4}{l}{\textbf{\textsc{En-Bridge}} (\textbf{X$=$en})} \\
encoder states (Tab. \ref{tab:pretrain_wmt} row $(6.3)$) &
26.8      & 26.3      & 15.2      & 20.9 \\
cluster centers & 
24.3      & 24.6      & 13.9      & 19.2 \\
\bottomrule
\end{tabular}
\caption{
\label{tab:cluster-center} 
At inference time, using cluster centers instead of the clustered encoder states 
degrades performance by $1.7$-$4.1$ spBLEU.
Despite the degradation, the scores show that translation from the clusters centers is still meaningful. 
This also rules out the possibility of the learned codes collapsing to trivial repetitions.
}
\end{table}

%% file: tables/oracle_zs.tex
\begin{table}[t]
\small
\centering
\setlength\tabcolsep{3pt} 
\begin{tabular}{ llcccccc } 
\toprule
 \textbf{Oracle Bitext} & 
 \multicolumn{4}{c}{\textbf{Avg. spBLEU}$(\uparrow)$} \\ 
 \cmidrule{2-5} 
 &
 \textbf{$\rightarrow$X} & 
 \textbf{X$\rightarrow$} & 
 \textbf{Y$\leftrightarrow$Z} &
 \textbf{Avg.}\\
 \midrule
 \multicolumn{4}{l}{\textbf{\textsc{Id-Bridge}} (\textbf{X$=$id})} \\
best zero-shot (Tab. \ref{tab:pretrain_wmt} row $(3.2)$) &
27.5     & 24.1       & 18.4      & 22.1 \\
$1\%$ & 
26.9     & 23.9      & 18.4       & 21.9 \\
$10\%$      & 
27.3     & 24.5       & 19.7      & 22.8 \\
$100\%$ ($2.2$M bitext) & 
26.6     & 24.8       & 19.7      & 22.7 \\
\midrule
\multicolumn{4}{l}{\textbf{\textsc{En-Bridge}} (\textbf{X$=$en})} \\
best zero-shot (Tab. \ref{tab:pretrain_wmt} row $(6.1)$) &
27.9      & 27.2      & 17.3      & 22.4\\
$1\%$ & 
27.0      & 26.8      & 17.1      & 22.0 \\
$10\%$      & 
27.5      & 27.4      & 18.4      & 22.9 \\
$100\%$ ($2.2$M bitext) & 
27.7      & 27.5      & 18.9      & 23.2 \\
\bottomrule
\end{tabular}
\caption{
\label{tab:oracle-zs} 
Impact of adding oracle parallel data for the previously zero-shot directions.
Adding $10$\% parallel data (roughly $220$K sentence pairs in our case) surpasses the best performance on direct zero-shot translation.
}
\end{table}

%% file: tables/multi_bridge.tex
\begin{table}[t]
\small
\centering
\setlength\tabcolsep{2.2pt} 
\begin{tabular}{ llcccccc } 
\toprule
 \multicolumn{2}{l}{\textbf{Data Condition}} & 
 \multicolumn{4}{c}{\textbf{Avg. spBLEU}$(\uparrow)$} \\  
 \cmidrule{3-6} 
 &&
 \textbf{$\rightarrow$X} & 
 \textbf{X$\rightarrow$} & 
 \textbf{Y$\leftrightarrow$Z} &
 \textbf{Avg.}\\
 \midrule
 \multirow{ 2}{*}{\textsc{Multi-Bridge}}
 & 
 X= id &   27.0 &  24.3 & \multirow{ 2}{*}{18.3} & 21.9 \\
 & 
 X= en &   27.8 &  27.7 && 23.0 \\
 \midrule
 \multicolumn{2}{l}{
 Only \textsc{Id-Bridge} (Tab. \ref{tab:pretrain_wmt} row $(3)$)
 } & 
 27.1      & 24.2      & 17.7      & 21.7\\
 \multicolumn{2}{l}{
 Only \textsc{En-Bridge} (Tab. \ref{tab:pretrain_wmt} row $(6)$)
 } &
 28.1      & 27.6      & \hspace{4pt}5.1       & 16.5 \\
 \bottomrule
\end{tabular}
\caption{
\label{tab:multibridge} Results of using multiple bridges (combining \textsc{Id-Bridge} and \textsc{En-Bridge}). 
Despite substantial gains over \textsc{En-Bridge}, 
the multi-bridge system only gives a mild improvement in zero-shot performance ($Y$$\leftrightarrow$$Z$) over the \textsc{Id-Bridge} system.
}
\end{table}

%% file: emnlp2022.bbl
\begin{thebibliography}{67}
\expandafter\ifx\csname natexlab\endcsname\relax\def\natexlab#1{#1}\fi

\bibitem[{Abdelali et~al.(2014)Abdelali, Guzman, Sajjad, and
  Vogel}]{abdelali-etal-2014-amara}
Ahmed Abdelali, Francisco Guzman, Hassan Sajjad, and Stephan Vogel. 2014.
\newblock \href
  {http://www.lrec-conf.org/proceedings/lrec2014/pdf/877_Paper.pdf} {The
  {AMARA} corpus: Building parallel language resources for the educational
  domain}.
\newblock In \emph{Proceedings of the Ninth International Conference on
  Language Resources and Evaluation ({LREC}'14)}, pages 1856--1862, Reykjavik,
  Iceland. European Language Resources Association (ELRA).

\bibitem[{Aharoni et~al.(2019)Aharoni, Johnson, and
  Firat}]{aharoni-etal-2019-massively}
Roee Aharoni, Melvin Johnson, and Orhan Firat. 2019.
\newblock \href {https://doi.org/10.18653/v1/N19-1388} {Massively multilingual
  neural machine translation}.
\newblock In \emph{Proceedings of the 2019 Conference of the North {A}merican
  Chapter of the Association for Computational Linguistics: Human Language
  Technologies, Volume 1 (Long and Short Papers)}, pages 3874--3884,
  Minneapolis, Minnesota. Association for Computational Linguistics.

\bibitem[{Aji et~al.(2020)Aji, Bogoychev, Heafield, and
  Sennrich}]{aji-etal-2020-neural}
Alham~Fikri Aji, Nikolay Bogoychev, Kenneth Heafield, and Rico Sennrich. 2020.
\newblock \href {https://doi.org/10.18653/v1/2020.acl-main.688} {In neural
  machine translation, what does transfer learning transfer?}
\newblock In \emph{Proceedings of the 58th Annual Meeting of the Association
  for Computational Linguistics}, pages 7701--7710, Online. Association for
  Computational Linguistics.

\bibitem[{Akhbardeh et~al.(2021)Akhbardeh, Arkhangorodsky, Biesialska, Bojar,
  Chatterjee, Chaudhary, Costa-jussa, Espa{\~n}a-Bonet, Fan, Federmann,
  Freitag, Graham, Grundkiewicz, Haddow, Harter, Heafield, Homan, Huck,
  Amponsah-Kaakyire, Kasai, Khashabi, Knight, Kocmi, Koehn, Lourie, Monz,
  Morishita, Nagata, Nagesh, Nakazawa, Negri, Pal, Tapo, Turchi, Vydrin, and
  Zampieri}]{akhbardeh-etal-2021-findings}
Farhad Akhbardeh, Arkady Arkhangorodsky, Magdalena Biesialska, Ond{\v{r}}ej
  Bojar, Rajen Chatterjee, Vishrav Chaudhary, Marta~R. Costa-jussa, Cristina
  Espa{\~n}a-Bonet, Angela Fan, Christian Federmann, Markus Freitag, Yvette
  Graham, Roman Grundkiewicz, Barry Haddow, Leonie Harter, Kenneth Heafield,
  Christopher Homan, Matthias Huck, Kwabena Amponsah-Kaakyire, Jungo Kasai,
  Daniel Khashabi, Kevin Knight, Tom Kocmi, Philipp Koehn, Nicholas Lourie,
  Christof Monz, Makoto Morishita, Masaaki Nagata, Ajay Nagesh, Toshiaki
  Nakazawa, Matteo Negri, Santanu Pal, Allahsera~Auguste Tapo, Marco Turchi,
  Valentin Vydrin, and Marcos Zampieri. 2021.
\newblock \href {https://aclanthology.org/2021.wmt-1.1} {Findings of the 2021
  conference on machine translation ({WMT}21)}.
\newblock In \emph{Proceedings of the Sixth Conference on Machine Translation},
  pages 1--88, Online. Association for Computational Linguistics.

\bibitem[{Al-Shedivat and Parikh(2019)}]{al-shedivat-parikh-2019-consistency}
Maruan Al-Shedivat and Ankur Parikh. 2019.
\newblock \href {https://doi.org/10.18653/v1/N19-1121} {Consistency by
  agreement in zero-shot neural machine translation}.
\newblock In \emph{Proceedings of the 2019 Conference of the North {A}merican
  Chapter of the Association for Computational Linguistics: Human Language
  Technologies, Volume 1 (Long and Short Papers)}, pages 1184--1197,
  Minneapolis, Minnesota. Association for Computational Linguistics.

\bibitem[{Anastasopoulos et~al.(2021)Anastasopoulos, Bojar, Bremerman, Cattoni,
  Elbayad, Federico, Ma, Nakamura, Negri, Niehues, Pino, Salesky, St{\"u}ker,
  Sudoh, Turchi, Waibel, Wang, and Wiesner}]{anastasopoulos-etal-2021-findings}
Antonios Anastasopoulos, Ond{\v{r}}ej Bojar, Jacob Bremerman, Roldano Cattoni,
  Maha Elbayad, Marcello Federico, Xutai Ma, Satoshi Nakamura, Matteo Negri,
  Jan Niehues, Juan Pino, Elizabeth Salesky, Sebastian St{\"u}ker, Katsuhito
  Sudoh, Marco Turchi, Alexander Waibel, Changhan Wang, and Matthew Wiesner.
  2021.
\newblock \href {https://doi.org/10.18653/v1/2021.iwslt-1.1} {{FINDINGS} {OF}
  {THE} {IWSLT} 2021 {EVALUATION} {CAMPAIGN}}.
\newblock In \emph{Proceedings of the 18th International Conference on Spoken
  Language Translation (IWSLT 2021)}, pages 1--29, Bangkok, Thailand (online).
  Association for Computational Linguistics.

\bibitem[{Anastasopoulos et~al.(2020)Anastasopoulos, Cattelan, Dou, Federico,
  Federmann, Genzel, Guzm{\'a}n, Hu, Hughes, Koehn, Lazar, Lewis, Neubig, Niu,
  {\"O}ktem, Paquin, Tang, and Tur}]{anastasopoulos-etal-2020-tico}
Antonios Anastasopoulos, Alessandro Cattelan, Zi-Yi Dou, Marcello Federico,
  Christian Federmann, Dmitriy Genzel, Franscisco Guzm{\'a}n, Junjie Hu,
  Macduff Hughes, Philipp Koehn, Rosie Lazar, Will Lewis, Graham Neubig,
  Mengmeng Niu, Alp {\"O}ktem, Eric Paquin, Grace Tang, and Sylwia Tur. 2020.
\newblock \href {https://doi.org/10.18653/v1/2020.nlpcovid19-2.5} {{TICO}-19:
  the translation initiative for {CO}vid-19}.
\newblock In \emph{Proceedings of the 1st Workshop on {NLP} for {COVID}-19
  (Part 2) at {EMNLP} 2020}, Online. Association for Computational Linguistics.

\bibitem[{Arivazhagan et~al.(2019{\natexlab{a}})Arivazhagan, Bapna, Firat,
  Aharoni, Johnson, and Macherey}]{arivazhagan2019missing}
Naveen Arivazhagan, Ankur Bapna, Orhan Firat, Roee Aharoni, Melvin Johnson, and
  Wolfgang Macherey. 2019{\natexlab{a}}.
\newblock \href {http://arxiv.org/abs/1903.07091} {The missing ingredient in
  zero-shot neural machine translation}.
\newblock \emph{CoRR}, abs/1903.07091.

\bibitem[{Arivazhagan et~al.(2019{\natexlab{b}})Arivazhagan, Bapna, Firat,
  Lepikhin, Johnson, Krikun, Chen, Cao, Foster, Cherry, Macherey, Chen, and
  Wu}]{arivazhagan2019massively}
Naveen Arivazhagan, Ankur Bapna, Orhan Firat, Dmitry Lepikhin, Melvin Johnson,
  Maxim Krikun, Mia~Xu Chen, Yuan Cao, George~F. Foster, Colin Cherry, Wolfgang
  Macherey, Zhifeng Chen, and Yonghui Wu. 2019{\natexlab{b}}.
\newblock \href {http://arxiv.org/abs/1907.05019} {Massively multilingual
  neural machine translation in the wild: Findings and challenges}.
\newblock \emph{CoRR}, abs/1907.05019.

\bibitem[{Baevski et~al.(2020)Baevski, Zhou, Mohamed, and
  Auli}]{DBLP:conf/nips/BaevskiZMA20}
Alexei Baevski, Yuhao Zhou, Abdelrahman Mohamed, and Michael Auli. 2020.
\newblock \href
  {https://proceedings.neurips.cc/paper/2020/hash/92d1e1eb1cd6f9fba3227870bb6d7f07-Abstract.html}
  {wav2vec 2.0: {A} framework for self-supervised learning of speech
  representations}.
\newblock In \emph{Advances in Neural Information Processing Systems 33: Annual
  Conference on Neural Information Processing Systems 2020, NeurIPS 2020,
  December 6-12, 2020, virtual}.

\bibitem[{Bapna and Firat(2019)}]{bapna-firat-2019-simple}
Ankur Bapna and Orhan Firat. 2019.
\newblock \href {https://doi.org/10.18653/v1/D19-1165} {Simple, scalable
  adaptation for neural machine translation}.
\newblock In \emph{Proceedings of the 2019 Conference on Empirical Methods in
  Natural Language Processing and the 9th International Joint Conference on
  Natural Language Processing (EMNLP-IJCNLP)}, pages 1538--1548, Hong Kong,
  China. Association for Computational Linguistics.

\bibitem[{Bengio et~al.(2013)Bengio, L{\'{e}}onard, and
  Courville}]{straightthrough}
Yoshua Bengio, Nicholas L{\'{e}}onard, and Aaron~C. Courville. 2013.
\newblock \href {http://arxiv.org/abs/1308.3432} {Estimating or propagating
  gradients through stochastic neurons for conditional computation}.
\newblock \emph{CoRR}, abs/1308.3432.

\bibitem[{Chiang et~al.(2022)Chiang, Chen, Yeh, and
  Neubig}]{chiang-etal-2022-breaking}
Ting-Rui Chiang, Yi-Pei Chen, Yi-Ting Yeh, and Graham Neubig. 2022.
\newblock \href {https://doi.org/10.18653/v1/2022.findings-acl.218} {Breaking
  down multilingual machine translation}.
\newblock In \emph{Findings of the Association for Computational Linguistics:
  ACL 2022}, pages 2766--2780, Dublin, Ireland. Association for Computational
  Linguistics.

\bibitem[{Christodoulopoulos and Steedman(2015)}]{bible-uedin}
Christos Christodoulopoulos and Mark Steedman. 2015.
\newblock \href {https://doi.org/10.1007/s10579-014-9287-y} {A massively
  parallel corpus: the bible in 100 languages}.
\newblock \emph{Lang. Resour. Evaluation}, 49(2):375--395.

\bibitem[{Dong et~al.(2015)Dong, Wu, He, Yu, and Wang}]{dong-etal-2015-multi}
Daxiang Dong, Hua Wu, Wei He, Dianhai Yu, and Haifeng Wang. 2015.
\newblock \href {https://doi.org/10.3115/v1/P15-1166} {Multi-task learning for
  multiple language translation}.
\newblock In \emph{Proceedings of the 53rd Annual Meeting of the Association
  for Computational Linguistics and the 7th International Joint Conference on
  Natural Language Processing (Volume 1: Long Papers)}, pages 1723--1732,
  Beijing, China. Association for Computational Linguistics.

\bibitem[{El-Kishky et~al.(2020)El-Kishky, Chaudhary, Guzm{\'a}n, and
  Koehn}]{el-kishky-etal-2020-ccaligned}
Ahmed El-Kishky, Vishrav Chaudhary, Francisco Guzm{\'a}n, and Philipp Koehn.
  2020.
\newblock \href {https://doi.org/10.18653/v1/2020.emnlp-main.480} {{CCA}ligned:
  A massive collection of cross-lingual web-document pairs}.
\newblock In \emph{Proceedings of the 2020 Conference on Empirical Methods in
  Natural Language Processing (EMNLP)}, pages 5960--5969, Online. Association
  for Computational Linguistics.

\bibitem[{ElNokrashy et~al.(2022)ElNokrashy, Hendy, Maher, Afify, and
  Awadalla}]{elnokrashy2022language}
Muhammad~N. ElNokrashy, Amr Hendy, Mohamed Maher, Mohamed Afify, and
  Hany~Hassan Awadalla. 2022.
\newblock \href {https://doi.org/10.48550/arXiv.2208.05852} {Language tokens:
  {A} frustratingly simple approach improves zero-shot performance of
  multilingual translation}.
\newblock \emph{CoRR}, abs/2208.05852.

\bibitem[{Escolano et~al.(2021)Escolano, Costa-Jussà, and
  Fonollosa}]{https://doi.org/10.1002/asi.24395}
Carlos Escolano, Marta~R. Costa-Jussà, and José A.~R. Fonollosa. 2021.
\newblock \href {https://doi.org/https://doi.org/10.1002/asi.24395} {From
  bilingual to multilingual neural-based machine translation by incremental
  training}.
\newblock \emph{Journal of the Association for Information Science and
  Technology}, 72(2):190--203.

\bibitem[{Fan et~al.(2021)Fan, Bhosale, Schwenk, Ma, El{-}Kishky, Goyal,
  Baines, Celebi, Wenzek, Chaudhary, Goyal, Birch, Liptchinsky, Edunov, Auli,
  and Joulin}]{mm100}
Angela Fan, Shruti Bhosale, Holger Schwenk, Zhiyi Ma, Ahmed El{-}Kishky,
  Siddharth Goyal, Mandeep Baines, Onur Celebi, Guillaume Wenzek, Vishrav
  Chaudhary, Naman Goyal, Tom Birch, Vitaliy Liptchinsky, Sergey Edunov,
  Michael Auli, and Armand Joulin. 2021.
\newblock \href {http://jmlr.org/papers/v22/20-1307.html} {Beyond
  english-centric multilingual machine translation}.
\newblock \emph{The Journal of Machine Learning Research}, 22:107:1--107:48.

\bibitem[{Firat et~al.(2016)Firat, Cho, and Bengio}]{firat-etal-2016-multi}
Orhan Firat, Kyunghyun Cho, and Yoshua Bengio. 2016.
\newblock \href {https://doi.org/10.18653/v1/N16-1101} {Multi-way, multilingual
  neural machine translation with a shared attention mechanism}.
\newblock In \emph{Proceedings of the 2016 Conference of the North {A}merican
  Chapter of the Association for Computational Linguistics: Human Language
  Technologies}, pages 866--875, San Diego, California. Association for
  Computational Linguistics.

\bibitem[{Freitag and Firat(2020)}]{freitag-firat-2020-complete}
Markus Freitag and Orhan Firat. 2020.
\newblock \href {https://aclanthology.org/2020.wmt-1.66} {Complete multilingual
  neural machine translation}.
\newblock In \emph{Proceedings of the Fifth Conference on Machine Translation},
  pages 550--560, Online. Association for Computational Linguistics.

\bibitem[{Ganin et~al.(2016)Ganin, Ustinova, Ajakan, Germain, Larochelle,
  Laviolette, Marchand, and Lempitsky}]{ganin2016domain}
Yaroslav Ganin, Evgeniya Ustinova, Hana Ajakan, Pascal Germain, Hugo
  Larochelle, Fran{\c{c}}ois Laviolette, Mario Marchand, and Victor Lempitsky.
  2016.
\newblock Domain-adversarial training of neural networks.
\newblock \emph{The Journal of Machine Learning Research}, 17(1):2096--2030.

\bibitem[{Goyal et~al.(2022)Goyal, Gao, Chaudhary, Chen, Wenzek, Ju, Krishnan,
  Ranzato, Guzmán, and Fan}]{10.1162/tacl_a_00474}
Naman Goyal, Cynthia Gao, Vishrav Chaudhary, Peng-Jen Chen, Guillaume Wenzek,
  Da~Ju, Sanjana Krishnan, Marc’Aurelio Ranzato, Francisco Guzmán, and
  Angela Fan. 2022.
\newblock \href {https://doi.org/10.1162/tacl_a_00474} {{The Flores-101
  Evaluation Benchmark for Low-Resource and Multilingual Machine Translation}}.
\newblock \emph{Transactions of the Association for Computational Linguistics},
  10:522--538.

\bibitem[{Gu et~al.(2019)Gu, Wang, Cho, and Li}]{gu-etal-2019-improved}
Jiatao Gu, Yong Wang, Kyunghyun Cho, and Victor~O.K. Li. 2019.
\newblock \href {https://doi.org/10.18653/v1/P19-1121} {Improved zero-shot
  neural machine translation via ignoring spurious correlations}.
\newblock In \emph{Proceedings of the 57th Annual Meeting of the Association
  for Computational Linguistics}, pages 1258--1268, Florence, Italy.
  Association for Computational Linguistics.

\bibitem[{Ha et~al.(2016)Ha, Niehues, and Waibel}]{ha-etal-2016-toward}
Thanh-Le Ha, Jan Niehues, and Alex Waibel. 2016.
\newblock \href {https://aclanthology.org/2016.iwslt-1.6} {Toward multilingual
  neural machine translation with universal encoder and decoder}.
\newblock In \emph{Proceedings of the 13th International Conference on Spoken
  Language Translation}, Seattle, Washington D.C. International Workshop on
  Spoken Language Translation.

\bibitem[{Johnson et~al.(2017)Johnson, Schuster, Le, Krikun, Wu, Chen, Thorat,
  Vi{\'e}gas, Wattenberg, Corrado, Hughes, and
  Dean}]{johnson-etal-2017-googles}
Melvin Johnson, Mike Schuster, Quoc~V. Le, Maxim Krikun, Yonghui Wu, Zhifeng
  Chen, Nikhil Thorat, Fernanda Vi{\'e}gas, Martin Wattenberg, Greg Corrado,
  Macduff Hughes, and Jeffrey Dean. 2017.
\newblock \href {https://doi.org/10.1162/tacl_a_00065} {{G}oogle{'}s
  multilingual neural machine translation system: Enabling zero-shot
  translation}.
\newblock \emph{Transactions of the Association for Computational Linguistics},
  5:339--351.

\bibitem[{Kaiser et~al.(2018)Kaiser, Bengio, Roy, Vaswani, Parmar, Uszkoreit,
  and Shazeer}]{kaiser}
Lukasz Kaiser, Samy Bengio, Aurko Roy, Ashish Vaswani, Niki Parmar, Jakob
  Uszkoreit, and Noam Shazeer. 2018.
\newblock \href {http://proceedings.mlr.press/v80/kaiser18a.html} {Fast
  decoding in sequence models using discrete latent variables}.
\newblock In \emph{Proceedings of the 35th International Conference on Machine
  Learning, {ICML} 2018, Stockholmsm{\"{a}}ssan, Stockholm, Sweden, July 10-15,
  2018}, volume~80 of \emph{Proceedings of Machine Learning Research}, pages
  2395--2404. {PMLR}.

\bibitem[{Kim et~al.(2021)Kim, Besacier, Nikoulina, and
  Schwab}]{kim-etal-2021-multilingual}
Zae~Myung Kim, Laurent Besacier, Vassilina Nikoulina, and Didier Schwab. 2021.
\newblock \href {https://doi.org/10.18653/v1/2021.findings-acl.250} {Do
  multilingual neural machine translation models contain language pair specific
  attention heads?}
\newblock In \emph{Findings of the Association for Computational Linguistics:
  ACL-IJCNLP 2021}, pages 2832--2841, Online. Association for Computational
  Linguistics.

\bibitem[{Knowles and Larkin(2021)}]{knowles-larkin-2021-nrc}
Rebecca Knowles and Samuel Larkin. 2021.
\newblock \href {https://aclanthology.org/2021.wmt-1.107} {{NRC}-{CNRC} systems
  for {U}pper {S}orbian-{G}erman and {L}ower {S}orbian-{G}erman machine
  translation 2021}.
\newblock In \emph{Proceedings of the Sixth Conference on Machine Translation},
  pages 999--1008, Online. Association for Computational Linguistics.

\bibitem[{Kudugunta et~al.(2019)Kudugunta, Bapna, Caswell, and
  Firat}]{kudugunta-etal-2019-investigating}
Sneha Kudugunta, Ankur Bapna, Isaac Caswell, and Orhan Firat. 2019.
\newblock \href {https://doi.org/10.18653/v1/D19-1167} {Investigating
  multilingual {NMT} representations at scale}.
\newblock In \emph{Proceedings of the 2019 Conference on Empirical Methods in
  Natural Language Processing and the 9th International Joint Conference on
  Natural Language Processing (EMNLP-IJCNLP)}, pages 1565--1575, Hong Kong,
  China. Association for Computational Linguistics.

\bibitem[{Liao et~al.(2021)Liao, Khadivi, and
  Hewavitharana}]{liao-etal-2021-back}
Baohao Liao, Shahram Khadivi, and Sanjika Hewavitharana. 2021.
\newblock \href {https://aclanthology.org/2021.wmt-1.50} {Back-translation for
  large-scale multilingual machine translation}.
\newblock In \emph{Proceedings of the Sixth Conference on Machine Translation},
  pages 418--424, Online. Association for Computational Linguistics.

\bibitem[{Libovick{\'y} and
  Fraser(2021{\natexlab{a}})}]{libovicky-fraser-2021-findings}
Jind{\v{r}}ich Libovick{\'y} and Alexander Fraser. 2021{\natexlab{a}}.
\newblock \href {https://aclanthology.org/2021.wmt-1.72} {Findings of the {WMT}
  2021 shared tasks in unsupervised {MT} and very low resource supervised
  {MT}}.
\newblock In \emph{Proceedings of the Sixth Conference on Machine Translation},
  pages 726--732, Online. Association for Computational Linguistics.

\bibitem[{Libovick{\'y} and
  Fraser(2021{\natexlab{b}})}]{libovicky-fraser-2021-lmu}
Jind{\v{r}}ich Libovick{\'y} and Alexander Fraser. 2021{\natexlab{b}}.
\newblock \href {https://aclanthology.org/2021.wmt-1.105} {The {LMU} {M}unich
  systems for the {WMT}21 unsupervised and very low-resource translation task}.
\newblock In \emph{Proceedings of the Sixth Conference on Machine Translation},
  pages 989--994, Online. Association for Computational Linguistics.

\bibitem[{Liu and Niehues(2021)}]{liu-niehues-2021-maastricht}
Danni Liu and Jan Niehues. 2021.
\newblock \href {https://doi.org/10.18653/v1/2021.iwslt-1.15} {Maastricht
  university{'}s multilingual speech translation system for {IWSLT} 2021}.
\newblock In \emph{Proceedings of the 18th International Conference on Spoken
  Language Translation (IWSLT 2021)}, pages 138--143, Bangkok, Thailand
  (online). Association for Computational Linguistics.

\bibitem[{Liu et~al.(2021{\natexlab{a}})Liu, Niehues, Cross, Guzm{\'a}n, and
  Li}]{liu-etal-2021-improving-zero}
Danni Liu, Jan Niehues, James Cross, Francisco Guzm{\'a}n, and Xian Li.
  2021{\natexlab{a}}.
\newblock \href {https://doi.org/10.18653/v1/2021.acl-long.101} {Improving
  zero-shot translation by disentangling positional information}.
\newblock In \emph{Proceedings of the 59th Annual Meeting of the Association
  for Computational Linguistics and the 11th International Joint Conference on
  Natural Language Processing (Volume 1: Long Papers)}, pages 1259--1273,
  Online. Association for Computational Linguistics.

\bibitem[{Liu et~al.(2021{\natexlab{b}})Liu, Zhang, Yin, and
  Zhu}]{liu-etal-2021-improving}
Hui Liu, Danqing Zhang, Bing Yin, and Xiaodan Zhu. 2021{\natexlab{b}}.
\newblock \href {https://doi.org/10.18653/v1/2021.naacl-main.83} {Improving
  pretrained models for zero-shot multi-label text classification through
  reinforced label hierarchy reasoning}.
\newblock In \emph{Proceedings of the 2021 Conference of the North American
  Chapter of the Association for Computational Linguistics: Human Language
  Technologies}, pages 1051--1062, Online. Association for Computational
  Linguistics.

\bibitem[{Mueller et~al.(2020)Mueller, Nicolai, McCarthy, Lewis, Wu, and
  Yarowsky}]{mueller-etal-2020-analysis}
Aaron Mueller, Garrett Nicolai, Arya~D. McCarthy, Dylan Lewis, Winston Wu, and
  David Yarowsky. 2020.
\newblock \href {https://aclanthology.org/2020.lrec-1.458} {An analysis of
  massively multilingual neural machine translation for low-resource
  languages}.
\newblock In \emph{Proceedings of the 12th Language Resources and Evaluation
  Conference}, pages 3710--3718, Marseille, France. European Language Resources
  Association.

\bibitem[{{NLLB Team} et~al.(2022){NLLB Team}, Costa-jussà, Cross, Çelebi,
  Elbayad, Heafield, Heffernan, Kalbassi, Lam, Licht, Maillard, Sun, Wang,
  Wenzek, Youngblood, Akula, Barrault, Gonzalez, Hansanti, Hoffman, Jarrett,
  Sadagopan, Rowe, Spruit, Tran, Andrews, Ayan, Bhosale, Edunov, Fan, Gao,
  Goswami, Guzmán, Koehn, Mourachko, Ropers, Saleem, Schwenk, and Wang}]{nllb}
{NLLB Team}, Marta~R. Costa-jussà, James Cross, Onur Çelebi, Maha Elbayad,
  Kenneth Heafield, Kevin Heffernan, Elahe Kalbassi, Janice Lam, Daniel Licht,
  Jean Maillard, Anna Sun, Skyler Wang, Guillaume Wenzek, Al~Youngblood, Bapi
  Akula, Loic Barrault, Gabriel~Mejia Gonzalez, Prangthip Hansanti, John
  Hoffman, Semarley Jarrett, Kaushik~Ram Sadagopan, Dirk Rowe, Shannon Spruit,
  Chau Tran, Pierre Andrews, Necip~Fazil Ayan, Shruti Bhosale, Sergey Edunov,
  Angela Fan, Cynthia Gao, Vedanuj Goswami, Francisco Guzmán, Philipp Koehn,
  Alexandre Mourachko, Christophe Ropers, Safiyyah Saleem, Holger Schwenk, and
  Jeff Wang. 2022.
\newblock \href {https://doi.org/10.48550/ARXIV.2207.04672} {No language left
  behind: Scaling human-centered machine translation}.

\bibitem[{Norouzi et~al.(2014)Norouzi, Mikolov, Bengio, Singer, Shlens, Frome,
  Corrado, and Dean}]{zs_ml_4}
Mohammad Norouzi, Tom{\'{a}}s Mikolov, Samy Bengio, Yoram Singer, Jonathon
  Shlens, Andrea Frome, Greg Corrado, and Jeffrey Dean. 2014.
\newblock \href {http://arxiv.org/abs/1312.5650} {Zero-shot learning by convex
  combination of semantic embeddings}.
\newblock In \emph{2nd International Conference on Learning Representations,
  {ICLR} 2014, Banff, AB, Canada, April 14-16, 2014, Conference Track
  Proceedings}.

\bibitem[{Ott et~al.(2019)Ott, Edunov, Baevski, Fan, Gross, Ng, Grangier, and
  Auli}]{ott-etal-2019-fairseq}
Myle Ott, Sergey Edunov, Alexei Baevski, Angela Fan, Sam Gross, Nathan Ng,
  David Grangier, and Michael Auli. 2019.
\newblock \href {https://doi.org/10.18653/v1/N19-4009} {fairseq: A fast,
  extensible toolkit for sequence modeling}.
\newblock In \emph{Proceedings of the 2019 Conference of the North {A}merican
  Chapter of the Association for Computational Linguistics (Demonstrations)},
  pages 48--53, Minneapolis, Minnesota. Association for Computational
  Linguistics.

\bibitem[{Pham et~al.(2021)Pham, Nguyen, Ha, St{\"u}ker, Waibel, and
  He}]{pham-etal-2021-multilingual}
Ngoc-Quan Pham, Tuan~Nam Nguyen, Thanh-Le Ha, Sebastian St{\"u}ker, Alexander
  Waibel, and Dan He. 2021.
\newblock \href {https://doi.org/10.18653/v1/2021.iwslt-1.18} {Multilingual
  speech translation {KIT} @ {IWSLT}2021}.
\newblock In \emph{Proceedings of the 18th International Conference on Spoken
  Language Translation (IWSLT 2021)}, pages 154--159, Bangkok, Thailand
  (online). Association for Computational Linguistics.

\bibitem[{Pham et~al.(2019)Pham, Niehues, Ha, and
  Waibel}]{pham-etal-2019-improving}
Ngoc-Quan Pham, Jan Niehues, Thanh-Le Ha, and Alexander Waibel. 2019.
\newblock \href {https://doi.org/10.18653/v1/W19-5202} {Improving zero-shot
  translation with language-independent constraints}.
\newblock In \emph{Proceedings of the Fourth Conference on Machine Translation
  (Volume 1: Research Papers)}, pages 13--23, Florence, Italy. Association for
  Computational Linguistics.

\bibitem[{Philip et~al.(2020)Philip, Berard, Gall{\'e}, and
  Besacier}]{philip-etal-2020-monolingual}
Jerin Philip, Alexandre Berard, Matthias Gall{\'e}, and Laurent Besacier. 2020.
\newblock \href {https://doi.org/10.18653/v1/2020.emnlp-main.361} {Monolingual
  adapters for zero-shot neural machine translation}.
\newblock In \emph{Proceedings of the 2020 Conference on Empirical Methods in
  Natural Language Processing (EMNLP)}, pages 4465--4470, Online. Association
  for Computational Linguistics.

\bibitem[{Popovi{\'c}(2017)}]{popovic-2017-chrf}
Maja Popovi{\'c}. 2017.
\newblock \href {https://doi.org/10.18653/v1/W17-4770} {chr{F}++: words helping
  character n-grams}.
\newblock In \emph{Proceedings of the Second Conference on Machine
  Translation}, pages 612--618, Copenhagen, Denmark. Association for
  Computational Linguistics.

\bibitem[{Raganato et~al.(2021)Raganato, V{\'a}zquez, Creutz, and
  Tiedemann}]{raganato-etal-2021-empirical}
Alessandro Raganato, Ra{\'u}l V{\'a}zquez, Mathias Creutz, and J{\"o}rg
  Tiedemann. 2021.
\newblock \href {https://doi.org/10.18653/v1/2021.emnlp-main.664} {An empirical
  investigation of word alignment supervision for zero-shot multilingual neural
  machine translation}.
\newblock In \emph{Proceedings of the 2021 Conference on Empirical Methods in
  Natural Language Processing}, pages 8449--8456, Online and Punta Cana,
  Dominican Republic. Association for Computational Linguistics.

\bibitem[{Rios et~al.(2020)Rios, M{\"u}ller, and
  Sennrich}]{rios-etal-2020-subword}
Annette Rios, Mathias M{\"u}ller, and Rico Sennrich. 2020.
\newblock \href {https://aclanthology.org/2020.wmt-1.64} {Subword segmentation
  and a single bridge language affect zero-shot neural machine translation}.
\newblock In \emph{Proceedings of the Fifth Conference on Machine Translation},
  pages 528--537, Online. Association for Computational Linguistics.

\bibitem[{Romera{-}Paredes and Torr(2015)}]{zs_ml_3}
Bernardino Romera{-}Paredes and Philip H.~S. Torr. 2015.
\newblock \href {http://proceedings.mlr.press/v37/romera-paredes15.html} {An
  embarrassingly simple approach to zero-shot learning}.
\newblock In \emph{Proceedings of the 32nd International Conference on Machine
  Learning, {ICML} 2015, Lille, France, 6-11 July 2015}, volume~37 of
  \emph{{JMLR} Workshop and Conference Proceedings}, pages 2152--2161.
  JMLR.org.

\bibitem[{Schwenk et~al.(2021)Schwenk, Chaudhary, Sun, Gong, and
  Guzm{\'a}n}]{schwenk-etal-2021-wikimatrix}
Holger Schwenk, Vishrav Chaudhary, Shuo Sun, Hongyu Gong, and Francisco
  Guzm{\'a}n. 2021.
\newblock \href {https://doi.org/10.18653/v1/2021.eacl-main.115}
  {{W}iki{M}atrix: Mining 135{M} parallel sentences in 1620 language pairs from
  {W}ikipedia}.
\newblock In \emph{Proceedings of the 16th Conference of the European Chapter
  of the Association for Computational Linguistics: Main Volume}, pages
  1351--1361, Online. Association for Computational Linguistics.

\bibitem[{Sennrich et~al.(2016)Sennrich, Haddow, and
  Birch}]{sennrich-etal-2016-improving}
Rico Sennrich, Barry Haddow, and Alexandra Birch. 2016.
\newblock \href {https://doi.org/10.18653/v1/P16-1009} {Improving neural
  machine translation models with monolingual data}.
\newblock In \emph{Proceedings of the 54th Annual Meeting of the Association
  for Computational Linguistics (Volume 1: Long Papers)}, pages 86--96, Berlin,
  Germany. Association for Computational Linguistics.

\bibitem[{Shazeer et~al.(2017)Shazeer, Mirhoseini, Maziarz, Davis, Le, Hinton,
  and Dean}]{moe}
Noam Shazeer, Azalia Mirhoseini, Krzysztof Maziarz, Andy Davis, Quoc~V. Le,
  Geoffrey~E. Hinton, and Jeff Dean. 2017.
\newblock \href {https://openreview.net/forum?id=B1ckMDqlg} {Outrageously large
  neural networks: The sparsely-gated mixture-of-experts layer}.
\newblock In \emph{5th International Conference on Learning Representations,
  {ICLR} 2017, Toulon, France, April 24-26, 2017, Conference Track
  Proceedings}. OpenReview.net.

\bibitem[{Socher et~al.(2013)Socher, Ganjoo, Manning, and Ng}]{zs_ml}
Richard Socher, Milind Ganjoo, Christopher~D. Manning, and Andrew~Y. Ng. 2013.
\newblock \href
  {https://proceedings.neurips.cc/paper/2013/hash/2d6cc4b2d139a53512fb8cbb3086ae2e-Abstract.html}
  {Zero-shot learning through cross-modal transfer}.
\newblock In \emph{Advances in Neural Information Processing Systems 26: 27th
  Annual Conference on Neural Information Processing Systems 2013. Proceedings
  of a meeting held December 5-8, 2013, Lake Tahoe, Nevada, United States},
  pages 935--943.

\bibitem[{Son and Lyu(2020)}]{son-lyu-2020-sparse}
Bokyung Son and Sungwon Lyu. 2020.
\newblock \href {https://doi.org/10.18653/v1/2020.findings-emnlp.205} {Sparse
  and decorrelated representations for stable zero-shot {NMT}}.
\newblock In \emph{Findings of the Association for Computational Linguistics:
  EMNLP 2020}, pages 2260--2266, Online. Association for Computational
  Linguistics.

\bibitem[{Tang et~al.(2021)Tang, Tran, Li, Chen, Goyal, Chaudhary, Gu, and
  Fan}]{tang-etal-2021-multilingual}
Yuqing Tang, Chau Tran, Xian Li, Peng-Jen Chen, Naman Goyal, Vishrav Chaudhary,
  Jiatao Gu, and Angela Fan. 2021.
\newblock \href {https://doi.org/10.18653/v1/2021.findings-acl.304}
  {Multilingual translation from denoising pre-training}.
\newblock In \emph{Findings of the Association for Computational Linguistics:
  ACL-IJCNLP 2021}, pages 3450--3466, Online. Association for Computational
  Linguistics.

\bibitem[{Tiedemann(2012)}]{tiedemann-2012-parallel}
J{\"o}rg Tiedemann. 2012.
\newblock \href
  {http://www.lrec-conf.org/proceedings/lrec2012/pdf/463_Paper.pdf} {Parallel
  data, tools and interfaces in {OPUS}}.
\newblock In \emph{Proceedings of the Eighth International Conference on
  Language Resources and Evaluation ({LREC}'12)}, pages 2214--2218, Istanbul,
  Turkey. European Language Resources Association (ELRA).

\bibitem[{Tjandra et~al.(2020)Tjandra, Sakti, and
  Nakamura}]{DBLP:conf/interspeech/TjandraS020}
Andros Tjandra, Sakriani Sakti, and Satoshi Nakamura. 2020.
\newblock \href {https://doi.org/10.21437/Interspeech.2020-3033} {Transformer
  {VQ-VAE} for unsupervised unit discovery and speech synthesis: Zerospeech
  2020 challenge}.
\newblock In \emph{Interspeech 2020, 21st Annual Conference of the
  International Speech Communication Association, Virtual Event, Shanghai,
  China, 25-29 October 2020}, pages 4851--4855. {ISCA}.

\bibitem[{van~den Oord et~al.(2017)van~den Oord, Vinyals, and
  Kavukcuoglu}]{vq-vae}
A{\"{a}}ron van~den Oord, Oriol Vinyals, and Koray Kavukcuoglu. 2017.
\newblock \href
  {https://proceedings.neurips.cc/paper/2017/hash/7a98af17e63a0ac09ce2e96d03992fbc-Abstract.html}
  {Neural discrete representation learning}.
\newblock In \emph{Advances in Neural Information Processing Systems 30: Annual
  Conference on Neural Information Processing Systems 2017, December 4-9, 2017,
  Long Beach, CA, {USA}}, pages 6306--6315.

\bibitem[{Vaswani et~al.(2017)Vaswani, Shazeer, Parmar, Uszkoreit, Jones,
  Gomez, Kaiser, and Polosukhin}]{transformer}
Ashish Vaswani, Noam Shazeer, Niki Parmar, Jakob Uszkoreit, Llion Jones,
  Aidan~N Gomez, \L~ukasz Kaiser, and Illia Polosukhin. 2017.
\newblock \href
  {http://papers.nips.cc/paper/7181-attention-is-all-you-need.pdf} {Attention
  is all you need}.
\newblock In I.~Guyon, U.~V. Luxburg, S.~Bengio, H.~Wallach, R.~Fergus,
  S.~Vishwanathan, and R.~Garnett, editors, \emph{Advances in Neural
  Information Processing Systems 30}, pages 5998--6008. Curran Associates, Inc.

\bibitem[{Wenzek et~al.(2021)Wenzek, Chaudhary, Fan, Gomez, Goyal, Jain, Kiela,
  Thrush, and Guzm{\'a}n}]{wenzek-etal-2021-findings}
Guillaume Wenzek, Vishrav Chaudhary, Angela Fan, Sahir Gomez, Naman Goyal,
  Somya Jain, Douwe Kiela, Tristan Thrush, and Francisco Guzm{\'a}n. 2021.
\newblock \href {https://aclanthology.org/2021.wmt-1.2} {Findings of the {WMT}
  2021 shared task on large-scale multilingual machine translation}.
\newblock In \emph{Proceedings of the Sixth Conference on Machine Translation},
  pages 89--99, Online. Association for Computational Linguistics.

\bibitem[{Wu et~al.(2021)Wu, Cheng, Wang, and Li}]{wu-etal-2021-language}
Liwei Wu, Shanbo Cheng, Mingxuan Wang, and Lei Li. 2021.
\newblock \href {https://doi.org/10.18653/v1/2021.findings-acl.264} {Language
  tags matter for zero-shot neural machine translation}.
\newblock In \emph{Findings of the Association for Computational Linguistics:
  ACL-IJCNLP 2021}, pages 3001--3007, Online. Association for Computational
  Linguistics.

\bibitem[{Xian et~al.(2017)Xian, Schiele, and Akata}]{zs_ml_2}
Yongqin Xian, Bernt Schiele, and Zeynep Akata. 2017.
\newblock \href {https://doi.org/10.1109/CVPR.2017.328} {Zero-shot learning -
  the good, the bad and the ugly}.
\newblock In \emph{2017 {IEEE} Conference on Computer Vision and Pattern
  Recognition, {CVPR} 2017, Honolulu, HI, USA, July 21-26, 2017}, pages
  3077--3086. {IEEE} Computer Society.

\bibitem[{Xie et~al.(2021)Xie, Hu, Yang, Yu, and Ju}]{xie-etal-2021-tentrans}
Wanying Xie, Bojie Hu, Han Yang, Dong Yu, and Qi~Ju. 2021.
\newblock \href {https://aclanthology.org/2021.wmt-1.53} {{T}en{T}rans
  large-scale multilingual machine translation system for {WMT}21}.
\newblock In \emph{Proceedings of the Sixth Conference on Machine Translation},
  pages 439--445, Online. Association for Computational Linguistics.

\bibitem[{Yang et~al.(2021{\natexlab{a}})Yang, Ma, Huang, Zhang, Dong, Huang,
  Muzio, Singhal, Hassan, Song, and Wei}]{yang-etal-2021-multilingual-machine}
Jian Yang, Shuming Ma, Haoyang Huang, Dongdong Zhang, Li~Dong, Shaohan Huang,
  Alexandre Muzio, Saksham Singhal, Hany Hassan, Xia Song, and Furu Wei.
  2021{\natexlab{a}}.
\newblock \href {https://aclanthology.org/2021.wmt-1.54} {Multilingual machine
  translation systems from {M}icrosoft for {WMT}21 shared task}.
\newblock In \emph{Proceedings of the Sixth Conference on Machine Translation},
  pages 446--455, Online. Association for Computational Linguistics.

\bibitem[{Yang et~al.(2021{\natexlab{b}})Yang, Eriguchi, Muzio, Tadepalli, Lee,
  and Hassan}]{yang-etal-2021-improving-multilingual}
Yilin Yang, Akiko Eriguchi, Alexandre Muzio, Prasad Tadepalli, Stefan Lee, and
  Hany Hassan. 2021{\natexlab{b}}.
\newblock \href {https://doi.org/10.18653/v1/2021.emnlp-main.578} {Improving
  multilingual translation by representation and gradient regularization}.
\newblock In \emph{Proceedings of the 2021 Conference on Empirical Methods in
  Natural Language Processing}, pages 7266--7279, Online and Punta Cana,
  Dominican Republic. Association for Computational Linguistics.

\bibitem[{Zhang et~al.(2021)Zhang, Bapna, Sennrich, and Firat}]{share-or-not}
Biao Zhang, Ankur Bapna, Rico Sennrich, and Orhan Firat. 2021.
\newblock \href {https://openreview.net/forum?id=Wj4ODo0uyCF} {Share or not?
  learning to schedule language-specific capacity for multilingual
  translation}.
\newblock In \emph{9th International Conference on Learning Representations,
  {ICLR} 2021, Virtual Event, Austria, May 3-7, 2021}. OpenReview.net.

\bibitem[{Zhang and Sennrich(2021)}]{zhang-sennrich-2021-edinburghs}
Biao Zhang and Rico Sennrich. 2021.
\newblock \href {https://doi.org/10.18653/v1/2021.iwslt-1.19} {{E}dinburgh{'}s
  end-to-end multilingual speech translation system for {IWSLT} 2021}.
\newblock In \emph{Proceedings of the 18th International Conference on Spoken
  Language Translation (IWSLT 2021)}, pages 160--168, Bangkok, Thailand
  (online). Association for Computational Linguistics.

\bibitem[{Zhang et~al.(2020)Zhang, Williams, Titov, and
  Sennrich}]{zhang-etal-2020-improving}
Biao Zhang, Philip Williams, Ivan Titov, and Rico Sennrich. 2020.
\newblock \href {https://doi.org/10.18653/v1/2020.acl-main.148} {Improving
  massively multilingual neural machine translation and zero-shot translation}.
\newblock In \emph{Proceedings of the 58th Annual Meeting of the Association
  for Computational Linguistics}, pages 1628--1639, Online. Association for
  Computational Linguistics.

\bibitem[{Zhu et~al.(2020)Zhu, Yu, Cheng, and Luo}]{zhu-etal-2020-language}
Changfeng Zhu, Heng Yu, Shanbo Cheng, and Weihua Luo. 2020.
\newblock \href {https://doi.org/10.18653/v1/2020.acl-main.150} {Language-aware
  interlingua for multilingual neural machine translation}.
\newblock In \emph{Proceedings of the 58th Annual Meeting of the Association
  for Computational Linguistics}, pages 1650--1655, Online. Association for
  Computational Linguistics.

\end{thebibliography}
